\documentclass[10pt,twocolumn,letterpaper]{article}

\usepackage{cvpr}              %

\usepackage[dvipsnames]{xcolor}
\usepackage{epsfig}
\usepackage{graphicx}
\usepackage{amsmath}
\usepackage{amssymb}
\usepackage{caption}
\usepackage{duckuments}
\usepackage{booktabs}
\usepackage{array,multirow,graphicx}
\usepackage{times}
\usepackage{gensymb}

\makeatletter
\def\NAT@spacechar{}
\makeatother

\newcommand{\trans}[0]{\mathbf{t}}
\newcommand{\R}[0]{\mathbb{R}}

\newcommand{\ours}[0]{\emph{RelPose++}}

\newcommand{\parhead}[1]{\vspace{2mm}\noindent \textbf{{#1}}}
\newcommand{\parsub}[1]{\vspace{1mm}\noindent\emph{{#1}}}

\DeclareMathOperator*{\argmax}{arg\,max}
\DeclareMathOperator*{\argmin}{arg\,min}

\definecolor{cvprblue}{rgb}{0.21,0.49,0.74}
\usepackage[breaklinks=true,bookmarks=false,colorlinks=true,citecolor=cvprblue,pagebackref=true]{hyperref}
\usepackage[capitalize]{cleveref}
\crefname{section}{Sec.}{Secs.}
\Crefname{section}{Section}{Sections}
\Crefname{table}{Table}{Tables}
\crefname{table}{Tab.}{Tabs.}
\Crefname{equation}{Equation}{Equations}
\crefname{equation}{eq.}{eqs.}

\def\paperID{69} %
\def\confName{3DV\xspace}
\def\confYear{2024\xspace}

\title{RelPose++: Recovering 6D Poses from Sparse-view Observations}

\author{Amy Lin* \hspace{4mm} Jason Y. Zhang* \hspace{4mm} Deva Ramanan \hspace{4mm} Shubham Tulsiani \vspace{2mm}\\ 
Carnegie Mellon University\\
{\tt\small \{amylin2,deva\}@cs.cmu.edu, \{jasonyzhang,shubhtuls\}@cmu.edu}
\\ {\tt \small \href{https://amyxlase.github.io/relpose-plus-plus}{https://amyxlase.github.io/relpose-plus-plus}}
}

\begin{document}
\twocolumn[{%
\renewcommand\twocolumn[1][]{#1}%
\maketitle
\begin{center}
    \newcommand{\teaserwidth}{\textwidth}
    \vspace{-6mm}
    \centerline{
    \includegraphics[width=\teaserwidth]{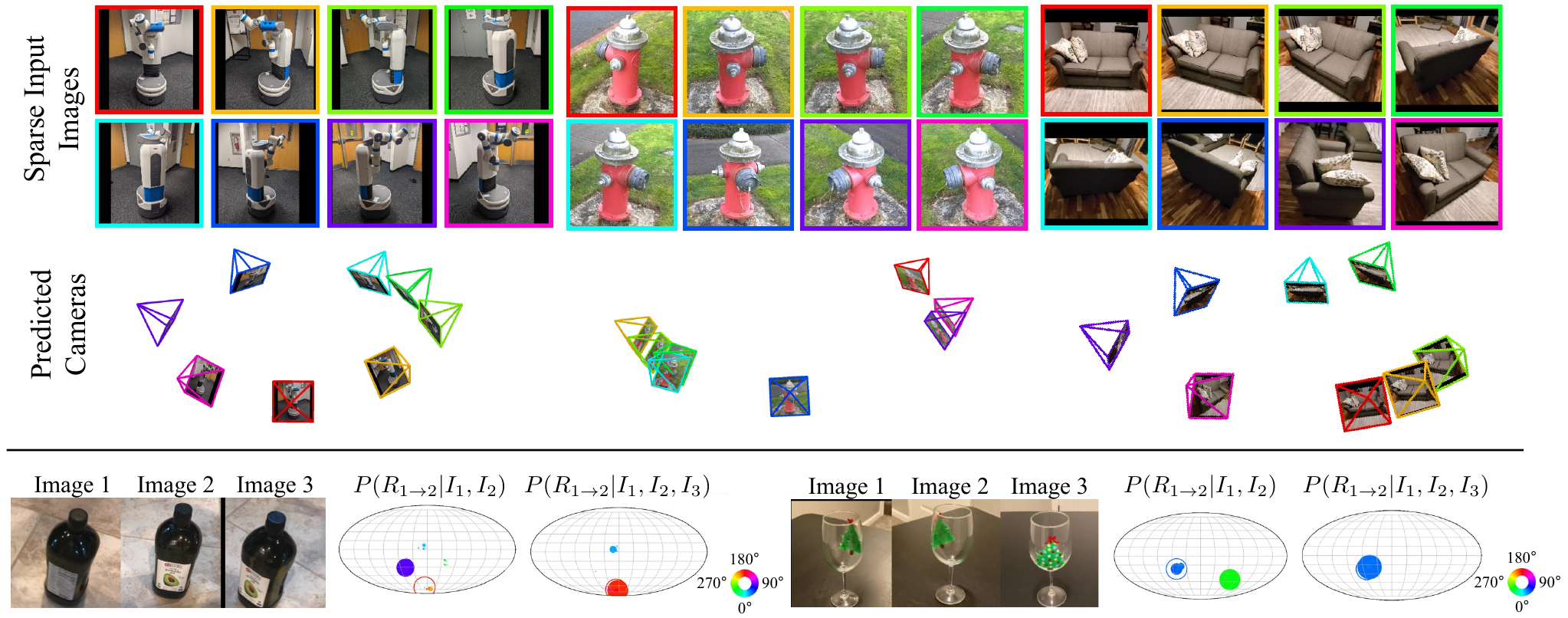}
     }
    \captionof{figure}{\textbf{Estimating 6D Camera Poses from Sparse Views.}
    We propose a framework \ours{} that, given a sparse set of input images, can infer the corresponding 6D camera rotations and translations ({\bf top}: the cameras are colored from red to magenta based on the image index). \ours{} estimates a probability distribution over the relative rotations of the cameras corresponding to any 2 images, but can do so while incorporating multi-view cues.
    We find that the distribution improves given additional images as context ({\bf bottom}).
    }
    \label{fig:teaser}
\end{center}%
}]

\begin{abstract}
\vspace{-4mm}
We address the task of estimating 6D camera poses from sparse-view image sets (2-8 images). This task is a vital pre-processing stage for nearly all contemporary (neural) reconstruction algorithms but remains challenging given sparse views, especially for objects with visual symmetries and texture-less surfaces. We build on the recent RelPose framework which learns a network that infers distributions over relative rotations over image pairs. We extend this approach in two key ways; first, we use attentional transformer layers to process multiple images jointly, since additional views of an object may resolve ambiguous symmetries in any given image pair (such as the handle of a mug that becomes visible in a third view). Second, we augment this network to also report camera translations by defining an appropriate coordinate system that decouples the ambiguity in rotation estimation from translation prediction. Our final system results in large improvements in 6D pose prediction over prior art on both seen and unseen object categories and also enables pose estimation and 3D reconstruction for in-the-wild objects.
\vspace{-6mm}
\end{abstract}
\section{Introduction}
\label{sec:intro}
The longstanding task of recovering 3D from 2D images has witnessed rapid progress over the recent years, with neural field-based methods~\cite{mildenhall2020nerf} enabling high-fidelity 3D capture of generic objects and scenes given dense multi-view observations.
There has also been a growing interest in enabling similar reconstructions in \emph{sparse-view} settings where only a few images of the underlying instance are available \eg online marketplaces, or casual captures by everyday users.
While several sparse-view reconstruction methods~\cite{yu2021pixelnerf,zhang2021ners,zhou2022sparsefusion} have shown promising results, they critically rely on known (precise or approximate) 6D camera poses for this 3D inference and sidestep the question of how these 6D poses can be acquired in the first place. In this work, we develop a system that can help bridge this gap and robustly infer (coarse) 6D poses given a sparse set of images for a generic object \eg a Fetch robot (\cref{fig:teaser}).

The classical approach~\cite{schoenberger2016sfm} to recovering camera poses from multiple images relies on bottom-up correspondences but is not robust in sparse-view settings with limited overlap across adjacent views. Our work instead adopts a top-down approach and builds on RelPose~\cite{zhang2022relpose}, which predicts distributions over pairwise relative rotations to then optimize multi-view consistent rotation hypotheses. While this optimization helps enforce multi-view consistency, RelPose's predicted distributions only consider pairs of images, which can be limiting. As an illustration, if we consider the first two images of the bottle shown in the bottom-left of \cref{fig:teaser}, we cannot narrow down the Y-axis rotation between the two (as the second label may be on the side or the back). However, if we consider the additional third image, we can immediately understand that the rotation between the first two should be nearly 180 degrees! %

We build on this insight in our proposed framework \ours{} and develop a method for jointly reasoning over multiple images for predicting pairwise relative distributions. Specifically, we incorporate a transformer-based module that leverages context across all input images to update the image-specific features subsequently used for relative rotation inference. \ours{} also goes beyond predicting only rotations and additionally infers the camera translation to yield 6D camera poses. A key hurdle is that the world coordinate frame used to define camera extrinsics can be arbitrary, and naive solutions to resolve this ambiguity (\eg instantiating the first camera as the world origin) end up entangling predictions of camera translations with predictions of (relative) camera rotations. Instead, for roughly center-facing images, we define a world coordinate frame centered at the intersection of cameras' optical axes. 
We show that this helps decouple the tasks of rotation and translation prediction, and leads to clear empirical gains.

\ours{} is trained on 41 categories from the CO3D dataset~\cite{reizenstein21co3d} and is able to recover 6D camera poses for objects from just a few images. We evaluate on seen categories, unseen categories, and even novel datasets (in a zero-shot fashion), improving rotation prediction by $10\%$ over prior art. We also evaluate the full 6D camera poses by measuring the accuracy of the predicted camera centers (while accounting for the similarity transform ambiguity), and demonstrate the benefits of our proposed coordinate system. We also formulate a metric that decouples the accuracy of predicted camera translations and predicted rotations, which may be generally useful for future benchmarking. Finally, we show that the  6D poses from \ours{} can be directly useful for downstream sparse-view 3D reconstruction methods.

\begin{figure*}[t]
    \centering
    \includegraphics[width=\textwidth]{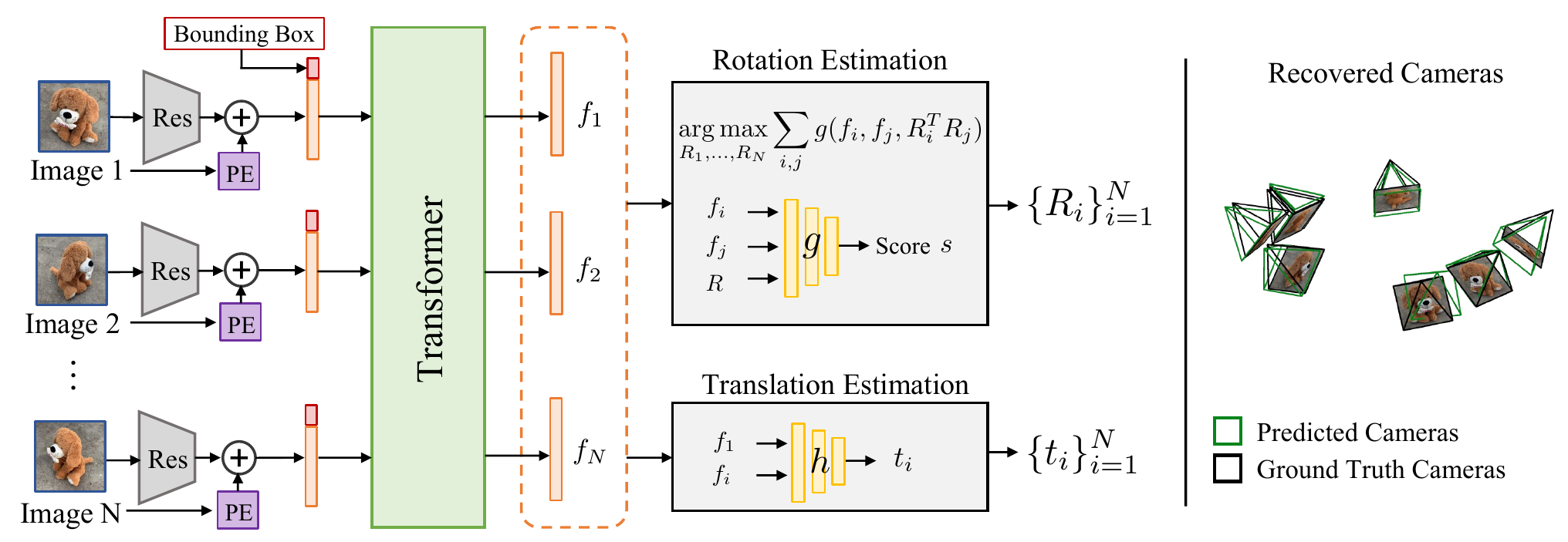}
    \caption{\textbf{Overview of RelPose++.} We present RelPose++, a method for sparse-view camera pose estimation. RelPose++ starts by extracting global image features using a ResNet 50. We positionally encode~\cite{vaswani2017attention} the image index and concatenate bounding box parameters as input to a Transformer.  %
    After processing all image features jointly, we separately estimate rotations and translations. To handle ambiguities in pose, we model the distribution of rotations using an energy-based formulation, following~\cite{implicitpdf2021,zhang2022relpose}. Because we predict the origin at the unique world coordinate closest to all optical axes, which is unambiguous (See \cref{sec:transpred} and \cref{fig:translation}), we can directly regress camera translation from the learned features. On the right, we visualize the recovered camera poses.}
    \label{fig:overview}
\end{figure*}
\section{Related Work}
\label{sec:related}

\vspace{-2mm}
\parhead{Pose Estimation Using Feature Correspondences.} The classic SfM and SLAM pipelines for pose estimation from sets of images or video streams involve computing matches~\cite{lucas1981iterative} between discriminative hand-crafted local features~\cite{bay2006surf,lowe2004distinctive}. These matches are used to estimate relative camera poses~\cite{longuet1981computer,nister2004efficient}, verified via RANSAC~\cite{fischler1981random}, and optimized via bundle adjustment~\cite{triggs1999bundle}. Subsequent research has explored improving each of these components. 
Learned feature estimation~\cite{detone2018superpoint} and feature matching~\cite{sarlin2020superglue,choy2016universal,liu2010sift} have improved robustness. 
This paradigm has been scaled by efficient parallelization~\cite{schoenberger2016mvs,schoenberger2016sfm} and can even run in real-time for visual odometry~\cite{mur2015orb,murORB2,ORBSLAM3_TRO}.
While we consider a similar task of estimating camera poses given images, our approach differs fundamentally because we do not rely on bottom-up correspondences as they cannot be reliably computed given sparse views.

\parhead{Single-view 3D Pose Estimation.}
In the extreme case of a single image, geometric cues are insufficient for reasoning about pose, so single-view 3D pose estimation approaches rely on learned data-driven priors. A significant challenge that arises in single-view 3D is that absolute pose must be defined with respect to a coordinate system. The typical solution is to assume a fixed set of categories (\eg humans~\cite{mehta2017vnect,KanazawaHMR17} or ShapeNet objects~\cite{chang2015shapenet}) with pre-defined canonical coordinate systems.
Related to our approach are methods that specifically handle object symmetries, which can be done by
predicting multiple hypotheses~\cite{manhardt2019explaining}, parameters for the antipodal Bingham distribution~\cite{okorn2020learning,gilitschenski2019deep}, or energy~\cite{implicitpdf2021} (similar to us). These methods predict absolute pose which is not well-defined without a canonical pose.

Because absolute poses only make sense in the context of a canonical pose, some single-view pose estimation papers have explored learning the canonical pose of objects automatically \cite{novotny2017learning,Xiao2020PoseContrast,sun2021canonical}. Other approaches bypass this issue by predicting poses conditioned on an input mesh~\cite{Xiao2019PoseFromShape,okorn2021zephyr,zhang2020phosa,wang2020NeMo} or point-cloud~\cite{wong2017segicp}.
In contrast, we resolve this issue by predicting relative poses from pairs of images.

\parhead{Learned Multi-view Pose Estimation.}
Given more images, it is still possible to learn a data-driven prior rather than rely on geometric consistency cues alone.
For instance, poses can directly be predicted using an RNN~\cite{wang2017deepvo,teed2021droid}, a transformer~\cite{nguyen2022pizza}, or auto-regressively~\cite{yang2020d3vo} for SLAM and object tracking applications. However, such approaches assume temporal locality not present in sparse-view images. Other approaches have incorporated category-specific priors, particularly for human pose~\cite{kanazawa2019learning,kocabas2020vibe,usman2022metapose,ma2022virtual}. In contrast, our work focuses on learning \textit{category-agnostic} priors that generalize beyond object categories seen at training.

Most related to our approach are methods that focus on sparse-view images. Such setups are more challenging since viewpoints have limited overlap.
In the case of using just 2 images for wide-baseline pose estimation, direct regression approaches~\cite{melekhov2017relative,rockwell20228} typically do not model uncertainty or require distributions to be Gaussian~\cite{chen2021wide}.
\cite{cai2021extreme} learns a 4D correlation volume from which distributions over relative rotations can be computed for pairs of patches.
Most similar to our work is the energy-based RelPose~\cite{zhang2022relpose}, which estimates distributions over relative rotations which can be composed together given more than 2 images. We build off of this energy-based framework and demonstrate significantly improved performance by incorporating multiview context. Additionally, RelPose only predicts rotations whereas we estimate 6D pose.

To estimate poses from sparse views, FORGE~\cite{jiang2022forge} and SparsePose~\cite{sinha23sparsepose} both directly regress 6D poses.
SparsePose also learns a bundle-adjustment procedure to refine predictions iteratively, but this refinement is complementary to our approach as it can improve any initial estimates.
Similarly, the concurrent PoseDiffusion~\cite{wang2023posediffusion} models a probabilistic bundle adjustment procedure via a diffusion model in contrast to the energy-based model in our work.

\section{Method}
Given a set of $N$ (roughly center-facing) input images \{$I_1, ..., I_N$\} of a generic object, we wish to recover consistent 6-DoF camera poses for each image \ie \{$(R_1, \trans_1), ..., (R_N, \trans_N)$\}, where $R_i$ and $\trans_i$ correspond to the rotation and translation for the $i^\text{th}$ camera viewpoint.

To estimate the camera rotations, we adopt the framework proposed by RelPose~\cite{zhang2022relpose}, where a consistent set of rotations can be obtained given pairwise relative rotation distributions (\cref{sec:relpose}). However, unlike RelPose which predicts these distributions using only two images, we incorporate a transformer-based module to allow the pairwise predicted distributions to capture multi-view cues (\cref{sec:mvrot}).  We then extend this multi-view reasoning module also to infer the translations associated with the cameras, while defining a world-coordinate system that helps reduce prediction ambiguity (\cref{sec:transpred}).

\subsection{Global Rotations from Pairwise Distributions}
\label{sec:relpose}
We build on RelPose~\cite{zhang2022relpose} for inferring consistent global rotations given a set of input images and briefly summarize the key components here. As absolute camera rotation prediction is ill-posed given the world-coordinate frame ambiguity, RelPose  infers pairwise relative rotations and then obtains a consistent set of global rotations. Using an energy-based model, it first approximates the (un-normalized) log-likelihood of the pairwise relative rotations given image features $f_i$ and $f_j$ with an MLP $g_{\theta}(f_i, f_j, R_{i \rightarrow j})$ which we treat as a negative energy or score.

Given the inferred distributions over pairwise relative rotations, RelPose casts the problem of finding global rotations as that of a mode-seeking optimization. Specifically, using a greedy initialization followed by block coordinate ascent, it recovers a set of global rotations that maximize the sum of relative rotation scores:
\begin{equation}
\label{eq:energy}
\{R_1,\ldots,R_N\} = \argmax_{\{R_i\}_{i=1}^N} \sum_{i,j} g_{\theta}( f_i, f_j,R_{i}^\top R_j)   
\end{equation}
In RelPose, the image features are extracted using a per-frame ResNet-50~\cite{he2016deep} encoder: $f_i=\varepsilon_\phi(I_i)$.

\subsection{Multi-view Cues for Pairwise Distributions}
\label{sec:mvrot}
Following RelPose, we similarly model the distribution of pairwise relative rotations using an energy-based model (\cref{eq:energy}).
However, instead of only relying on the images $I_i$ and $I_j$ to obtain the corresponding features $f_i$ and $f_j$, we propose a transformer-based module that allows for these features to depend on {\em other} images in the multi-view set.

\parhead{Multi-view Conditioned Image Features.} As illustrated in \cref{fig:overview}, we first use a ResNet~\cite{he2016deep} to extract per-image features.  We also add an ID-specific encoding to the ResNet features and concatenate an embedding of the bounding box used to obtain the input crop from the larger image (as it may be informative about the scene scale when inferring translation). Unlike RelPose which then directly feeds these image-specific features as input to the energy prediction module, we use a transformer (similar to other recent sparse-view works~\cite{wang2023posediffusion,nguyen2022pizza,sinha23sparsepose}) to update these features in the context of the other images.
We denote this combination of the feature extractor and transformer as a scene encoder $\mathcal{E}_{\phi}$, which given $N$ input images $\{I_n\}$ outputs multi-view conditioned features $\{f_n\}$ corresponding to each image:
\begin{equation}
    f_i = \mathcal{E}^i_{\phi}(I_1, \ldots I_N), \quad \forall i \in \{1 \ldots N\} %
\end{equation}

\parhead{Learning Objective.} Given a dataset with posed multi-view images of diverse objects, we  jointly train the scene encoder $\mathcal{E}_{\phi}$ and the pairwise energy-based model $g_{\theta}$ by simply minimizing the negative log-likelihood (NLL) of the true (relative) rotations~\cite{zhang2022relpose,implicitpdf2021}.  In particular, we randomly sample $N \in [2,8]$ images for a training object, and minimize the NLL of the true relative rotations $R_{i \rightarrow j}^{gt}$ under our predicted distribution:
\begin{equation}
    L_{\text{rot}} = \sum_{i,j} -\log~\frac{\exp{g_{\theta}(f_i, f_j, R_{i \rightarrow j}^{gt})}}{\sum_{R'} \exp{g_{\theta}(f_i, f_j, R')}}
\end{equation}

\subsection{Predicting Camera Translations}
\label{sec:transpred}

\begin{figure}
    \centering
    \includegraphics[width=\linewidth]{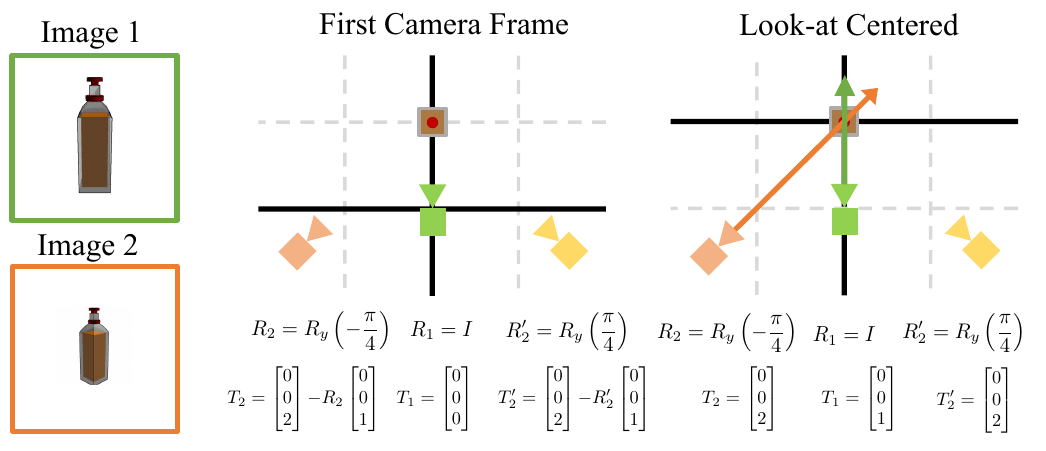}
    \caption{\textbf{Coordinate Systems for Estimating Camera Translation.}
    Given two images, consider the task of estimating their 6D poses, i.e., the $R$ and $T$ that transform points from the world frame to each camera's frame (\textbf{Left}).
    In typical SLAM setups, the world frame is centered at the first camera, but this implies
    the target camera translation $T_2$ depends on the target rotation $R_2$ (\textbf{Middle}).
    For symmetric objects where $R_2$ may be ambiguous, this may lead to unstable predictions for translation. Instead, for roughly center-facing cameras, a better solution is to set the world origin at the unique point closest to the optical axes of all cameras (\textbf{Right}).
    This helps decouple the task of predicting camera translations from rotations.
    }
    \label{fig:translation}
\end{figure}

\begin{figure*}
    \centering
    \includegraphics[width=\linewidth]{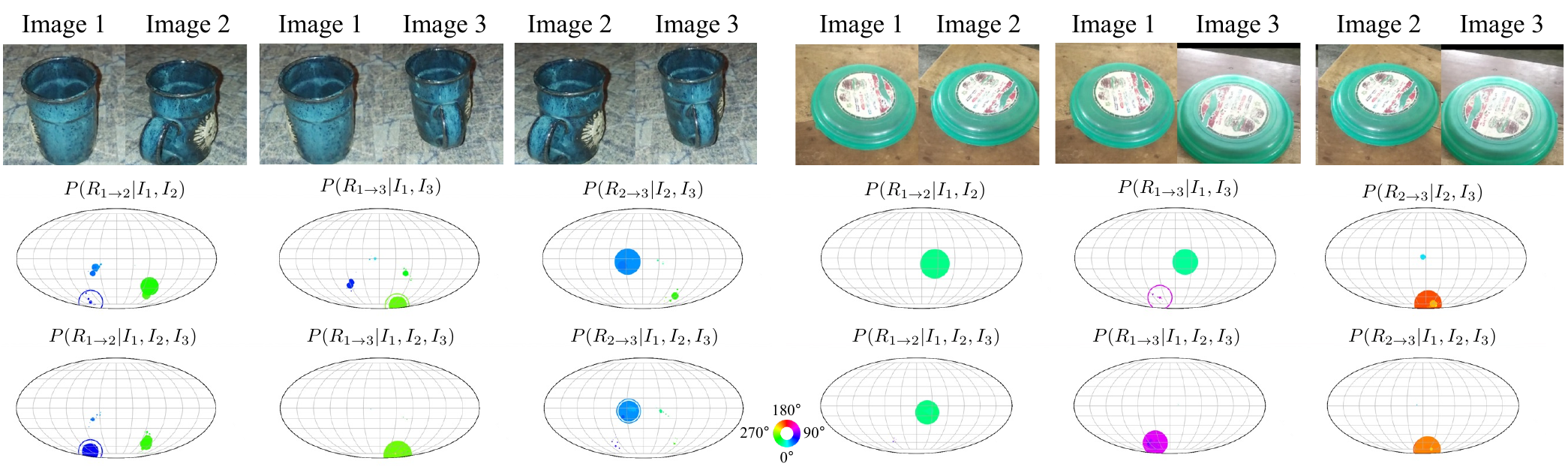}
    \caption{\textbf{Resolving Pose Ambiguity with More Images.} The relative rotation between only two views may be ambiguous for highly symmetric objects such as cups, frisbees, and apples. Often, seeing a third view will provide enough additional context to the scene to determine the correct relative rotation. When images are shown to the model in three separate pairs, as denoted by $P(R_{i\rightarrow j}|I_i, I_j)$, the output probability distribution may have more than one mode due to the symmetry of the object, but when shown all three images together to predict $P(R_{i\rightarrow j}|I_1, I_2, I_3)$, the model has a significantly more confident prediction. Following~\cite{zhang2022relpose}, we visualize distributions over relative rotations by projecting the rotation matrix such that the x-axis represents the yaw, the y-axis represents the pitch, and the color represents the roll. The size of each circle corresponds to probability, and rotations with negligible probability are filtered. The ground truth rotation is denoted by the unfilled circle.}
    \label{fig:resolve_ambiguity}
\end{figure*}

Using the multi-view aware image features $f_i$, we can directly predict the per-image camera translation $\trans_i = h_{\psi}(f_i)$. However, a central hurdle to learning such prediction is the inherent ambiguity in the world coordinate system. Specifically, the `ground-truth' cameras obtained from SfM are meaningful only up to an arbitrary similarity transform~\cite{hartley2003multiple}), and training our network to predict these can lead to incoherent training targets across each sequence. We therefore first need to define a consistent coordinate frame across training instances, so that the networks can learn to make meaningful predictions.

\parhead{Geometric Interpretation of Camera Translation.} Recall that the camera extrinsics $(R_i,\trans_i)$ define a transformation of a point $\mathbf{x}^w$ in world frame to camera frame $\mathbf{x}^c_i = R_i \mathbf{x}^w + \trans_i$. The translation $\trans_i$ is therefore the location of the world origin in each camera's coordinate frame (and not the location of the camera in the world frame!). We can also see that an arbitrary rotation of the world coordinate system ($\bar{\mathbf{x}}^w = \Delta R ~\mathbf{x}^w$), does not affect the per-camera translations and that only the location of the chosen world origin (and the scaling) are relevant factors. To define a consistent coordinate frame for predicting rotations, we must therefore decide where to place the world origin and how to choose an appropriate scale.

\parhead{Look-at Centered Coordinate System.} One convenient choice, often also adopted by SfM/SLAM approaches~\cite{schoenberger2016sfm,davison2007monoslam}, is to define the world coordinate system as centered on the first camera (denoted as `First Camera Frame'). Unfortunately, the per-camera translations in this coordinate frame entangle the relative rotations between cameras (as $\trans_i$ is the location of the first camera in the $i^{th}$ camera's frame). As illustrated in \cref{fig:translation}, ambiguity in estimating this relative transformation for (\eg, symmetric) objects can lead to uncertainty in the translation prediction.

Instead, we argue that for roughly center-facing captures, one should define the unique point closest to the optical axes of the input cameras as the world origin. 
Intuitively, this is akin to setting the `object center' as the world origin, and the translation then simply corresponds to the inference of where the object is in the camera frame (and this remains invariant even if one is unsure of camera rotation as illustrated in \cref{fig:translation}). However, instead of relying on a semantically defined `object center' which may be ambiguous given partial observations, the closest approach point across optical axes is a well-defined geometric proxy. Finally, to resolve scale ambiguity, we assume that the first camera is a unit distance away from this point. 

\parhead{Putting it Together.}
In addition to the energy-based predictor (Eq. \ref{eq:energy}), we also train a translation prediction module that infers the per-camera $\trans_i = h_{\psi}(f_1, f_i) \in \R^3$ given the multi-view features. Because we normalize the scene such that $\lVert \trans_1 \rVert = 1$, we provide $h_\psi$ with the first image feature $f_1$.
To define the target translations for training, we use the ground-truth cameras (SfM) $\{(R_i, \trans_i)\}$ to first identify the point $\mathbf{c}$ closest to all the optical axes. We can then transform the world coordinate to be centered at $\textbf{c}$, thus obtaining the target translations as $\bar{\trans}_i = s (\trans_i - R_i\mathbf{c})$, where the scale $s$ ensures a unit norm for $\bar{\trans}_1$.
For this training, we simply use an L1 loss between the target and predicted translations:
\begin{equation}
L_{\text{trans}} = \lVert h_{\psi}(f_i, f_1) - \bar{\trans}_i \rVert_1
\end{equation}
Together with the optimized global rotations, these predicted translations yield 6-DoF cameras given a sparse set of input images at inference.

\begin{figure*}
    \centering
    \includegraphics[width=\textwidth]{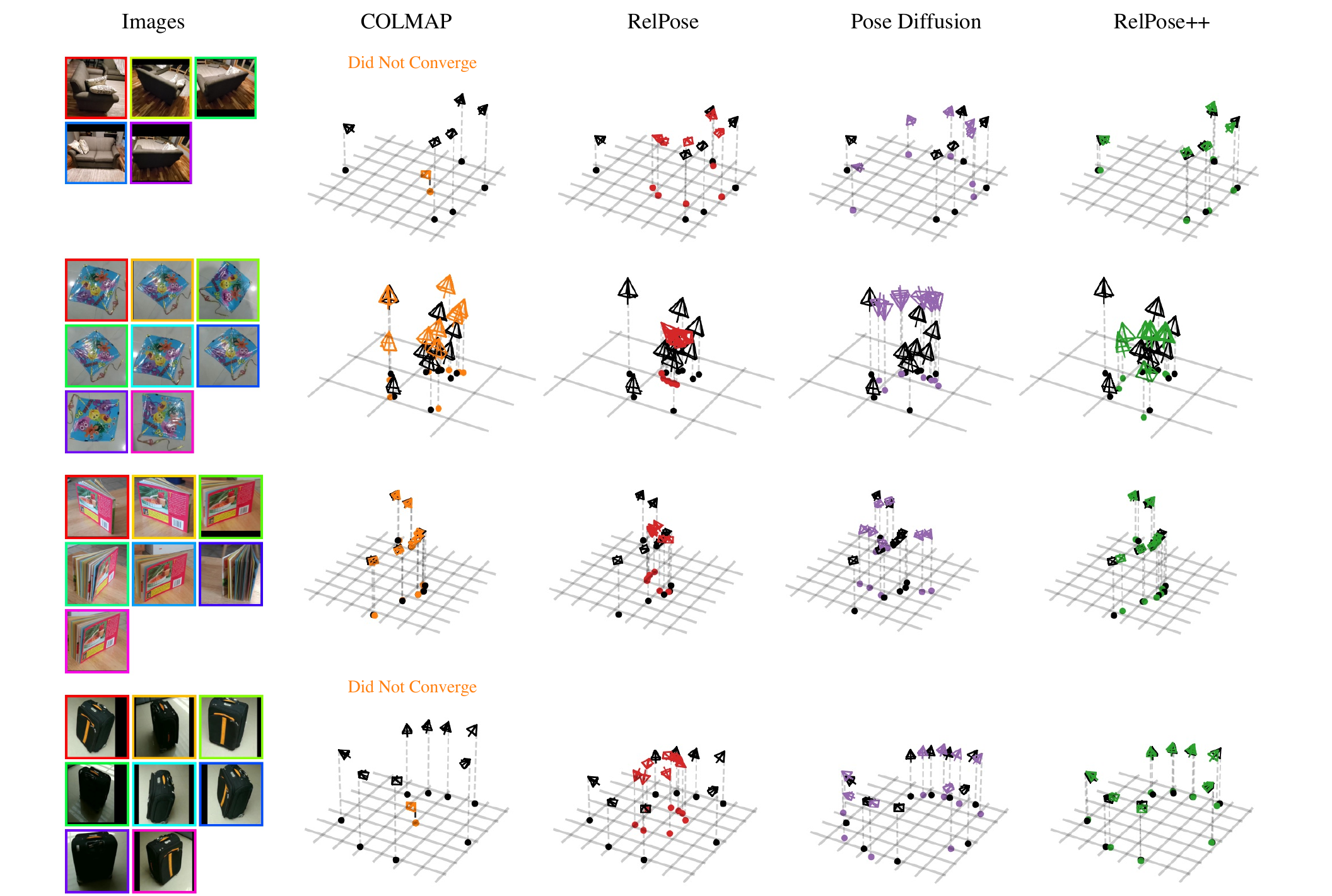}
    \caption{\textbf{Qualitative Results of Recovered Camera Trajectories.} We compare our approach with COLMAP, RelPose, and PoseDiffusion. Since RelPose does not predict translations, we set the translations to be unit distance from the scene center. We visualize predicted camera trajectories in color and the ground truth in black, aligned using a Procrustes optimal alignment on the camera centers.    
    We find that COLMAP is accurate but brittle, converging only occasionally when the object has highly discriminative features and sufficient overlap between images. RelPose, while mostly accurate, usually makes 1-2 mistakes per sequence which causes misalignment. PoseDiffusion is generally accurate but struggles sometimes with symmetry. We find that our method consistently outperforms the baselines.}
    \label{fig:qualitative}
\end{figure*}

\section{Evaluation}

\begin{table}
\begin{center}
\footnotesize
\resizebox{\linewidth}{!}{
\begin{tabular}{llccccccc}
\toprule
& \# of Images & 2 & 3 & 4 & 5 & 6 & 7 & 8\\
\midrule
\parbox[t]{2mm}{\multirow{6}{*}{\rotatebox[origin=c]{90}{Seen Cate.}}} & COLMAP (SP+SG)~\cite{sarlin2019coarse} & 30.7 &  28.4 &  26.5 &  26.8 &  27.0 &  28.1 &  30.6  \\
&  RelPose~\cite{zhang2022relpose} & 56.0 & 56.5 & 57.0 & 57.2 & 57.2 & 57.3 & 57.2 \\
& PoseDiffusion~\cite{wang2023posediffusion} & 75.2 & 76.6 & 77.0 & 77.3 & 77.7 & 78.2 & 78.5 \\
\cmidrule(lr){2-9}
& Pose Regression & 49.1 & 50.7 & 53.0 & 54.6 & 55.7 & 56.1 & 56.5 \\
& Ours (N=2) & \textbf{81.8} & 82.3 & 82.7 & 83.2 & 83.3 & 83.5 & 83.6\\
& Ours (Full)  & \textbf{81.8} & \textbf{82.8} & \textbf{84.1} & \textbf{84.7} & \textbf{84.9} & \textbf{85.3} & \textbf{85.5}\\
\midrule
\parbox[t]{2mm}{\multirow{6}{*}{\rotatebox[origin=c]{90}{Unseen Cate.}}}
& COLMAP (SP+SG)~\cite{sarlin2019coarse}& 34.5 &  31.8 &  31.0 &  31.7 &  32.7 &  35.0 &  38.5\\
& RelPose~\cite{zhang2022relpose} & 48.6 & 47.5 & 48.1 & 48.3 & 48.4 & 48.4 & 48.3 \\
& PoseDiffusion~\cite{wang2023posediffusion} & 60.0 & 64.8 & 64.6 & 65.8 & 65.7 & 66.6 & 67.8 \\
\cmidrule(lr){2-9}
& Pose Regression &  42.7 & 43.8 & 46.3 & 47.7 & 48.4 & 48.9 & 48.9\\
& Ours (N=2)  & \textbf{69.8} & 69.6 & 70.1 & 69.8 & 70.4 & 70.5 & 71.2\\
& Ours (Full)  &  \textbf{69.8} & \textbf{71.1} & \textbf{71.9} & \textbf{72.8} & \textbf{73.8} & \textbf{74.4} & \textbf{74.9}\\
\bottomrule
\end{tabular}
}
\end{center}
\caption{\textbf{Joint Rotation Accuracy @ 15\degree.} We measure the relative angular error between pairs of relative predicted and ground truth rotations . We report the proportion of angular errors within 15 degrees and report accuracies for varying thresholds in the supplement. 
With more images, our method surpasses the ablation that only looks at 2 images ($N$=2), showing the benefit of context.}
\label{tab:joint_rotations}
\end{table}

\begin{table}
\begin{center}
\footnotesize
\resizebox{\linewidth}{!}{
\begin{tabular}{llccccccc}
\toprule
& \# of Images & 2 & 3 & 4 & 5 & 6 & 7 & 8\\
\midrule
\parbox[t]{2mm}{\multirow{5}{*}{\rotatebox[origin=c]{90}{Seen Cate.}}} &  COLMAP (SP+SG)~\cite{sarlin2019coarse} & 100 & 35.8 & 26.1 & 21.6 & 18.9 & 18.3 & 19.2\\
& PoseDiffusion~\cite{wang2023posediffusion} & 100 & 86.6 & 80.5 & 77.2 & 75.9 & 74.4 & 73.7 \\
\cmidrule(lr){2-9}
& Pose Reg. (First Fr.) & 100 & 87.6 & 81.2 & 77.6 & 75.8 & 74.5 & 73.6 \\
& Pose Reg. (Our Fr.) & 100 & 90.3 & 84.6 & 81.5 & 80.0 & 78.5 & 77.7 \\
&  Ours & 100 & \textbf{92.3} & \textbf{89.1} & \textbf{87.5} & \textbf{86.3} & \textbf{85.9} & \textbf{85.5} \\
\midrule
\parbox[t]{2mm}{\multirow{5}{*}{\rotatebox[origin=c]{90}{Unseen Cate.}}} & COLMAP (SP+SG)~\cite{sarlin2019coarse} &  100 & 37.9 & 29.3 & 24.7 & 23.1 & 23.5 & 25.3\\
& PoseDiffusion~\cite{wang2023posediffusion} & 100 & 78.0 & 65.8 & 61.3 & 57.0 & 54.4 & 55.1 \\
\cmidrule(lr){2-9}
& Pose Reg. (First Fr.) & 100 & 78.8 & 71.4 & 66.3 & 63.6 & 61.8 & 60.4 \\
& Pose Reg. (Our Fr.) & 100 & \textbf{82.8} & 74.0 & 70.0 & 67.8 & 65.8 & 65.3 \\
&  Ours & 100 & 82.5 & \textbf{75.6} & \textbf{71.9} & \textbf{69.9} & \textbf{68.5} & \textbf{67.5}\\
\bottomrule
\end{tabular}}
\end{center}
\caption{\textbf{Camera Center Accuracy @ 0.2.} We report the proportion of camera centers that are within 20\% of the scene scale to the ground truth camera centers. We align the predicted and ground truth camera centers using an optimal 7-DoF similarity transform (hence all methods are at 100\% for $N$=2 and performance appears to drop with more images as there are more constraints). 
}   
\label{tab:camera_center}
\end{table}

\subsection{Experimental Setup}
\vspace{-2mm}

\parhead{Dataset.} We train and test our models on the CO3D~\cite{reizenstein21co3d} (v2) dataset, which consists of turntable-style video sequences across 51 object categories. Each video sequence is associated with ground truth camera poses acquired using COLMAP~\cite{schoenberger2016sfm}.
Following~\cite{zhang2022relpose}, we train on 41 object categories and hold out the same 10 object categories to evaluate generalization. After filtering for the camera pose score, we train on a total of 22,375 sequences with 2,212,952 images.

\parhead{Task and Metrics.} We randomly sample 2 $\leq N \leq$ 8 center-cropped images $\{I_n\}$ from each test sequence. Given these as input, each approach then infers a set of global 6-DoF camera poses $\{R_i, \trans_i\}$ corresponding to each input image. To evaluate these predictions, we report accuracy under various complementary metrics, all of which are invariant under global similarity transforms for the prediction/ground-truth cameras. To reduce variance in metrics, we re-sample the $N$ images from each test sequence 5 times and compute the mean.

\parsub{Rotation Accuracy.} We evaluate relative rotation error between every pair of predicted and ground truth rotations. Following~\cite{zhang2022relpose,sinha23sparsepose}, we report the proportion of pose errors less than 15 degrees. 

\parsub{Camera Center Accuracy.} 
Following standard benchmarks in SLAM~\cite{sturm2012benchmark} that evaluated recovered poses using camera localization error, we measure the accuracy of the predicted camera centers. However, as the predicted centers $\mathbf{c}_i = -R_i\trans_i$ may be in a different coordinate system from the SfM camera centers $\mathbf{c}_i^{gt}$, we first compute the optimal similarity transform to align the predicted centers with the ground-truth ~\cite{umeyama1991least}. Following~\cite{sinha23sparsepose}, we then report the proportion of predicted camera centers within 20\% of the scale of the scene in~\cref{tab:camera_center}, where the scale is defined as the distance from the centroid of the ground truth camera centers to the furthest camera center.

\parhead{Baselines.} We compare our approach with state-of-the-art correspondence-based and learning-based methods:

\parsub{COLMAP~(SP+SG)~\cite{schoenberger2016mvs,schoenberger2016sfm}}. This represents a state-of-the-art
SFM pipeline (COLMAP) that uses SuperPoint features~\cite{detone2018superpoint} with SuperGlue matching~\cite{sarlin2020superglue}. We use the implementation provided by HLOC~\cite{sarlin2019coarse}. 

\parsub{RelPose~\cite{zhang2022relpose}.} We evaluate RelPose, which also uses a pairwise energy-based scoring network. As this only predicts rotations, we exclude it from translation evaluation. 

\parsub{PoseDiffusion~\cite{wang2023posediffusion}.} PoseDiffusion is a concurrent work that combines diffusion with geometric constraints (from correspondences) to infer sparse-view poses probabilistically. All evaluations are with the geometry-guided sampling.

\begin{figure*}
    \centering
    \includegraphics[width=\linewidth]{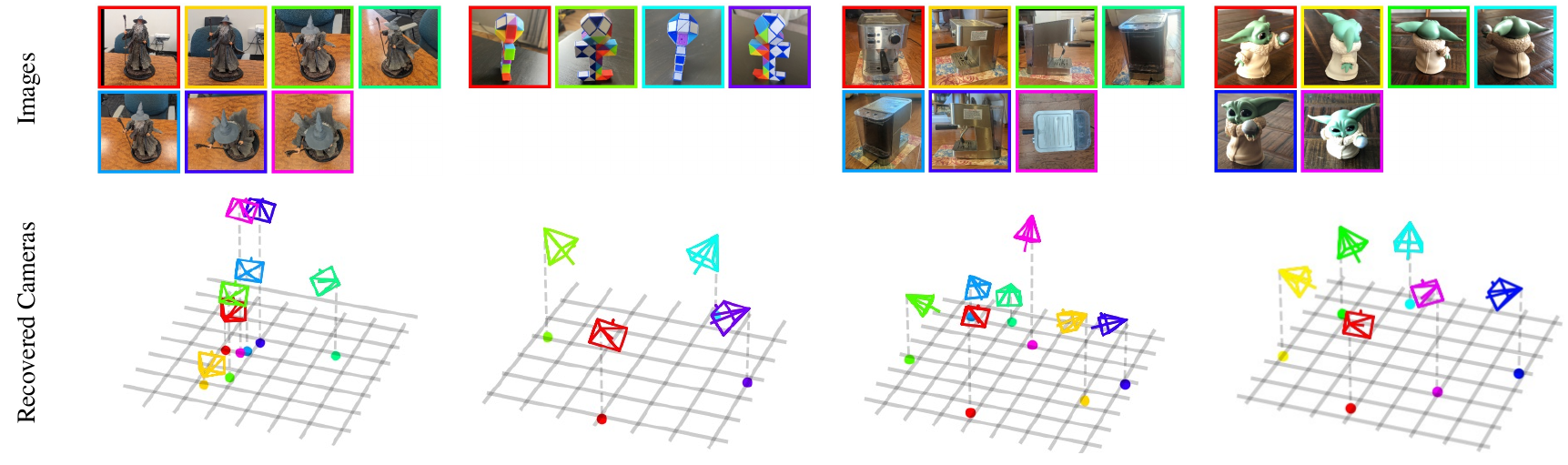}
    \caption{\textbf{Recovered Camera Poses from In-the-Wild Images.} We find that RelPose++ generalizes well to images outside of the distribution of CO3D object categories. Here, we demonstrate that RelPose++ can recover accurate camera poses even for self-captures of Gandalf the Grey, a Rubrik snake, an espresso machine, and Grogu. RelPose++ can capture challenging rotations and translations, including top-down poses, varying distances from the camera, and in-plane rotations (see Gandalf).}
    \label{fig:unconstrained_objects}
\end{figure*}

\begin{figure}
    \centering
    \includegraphics[width=\linewidth]{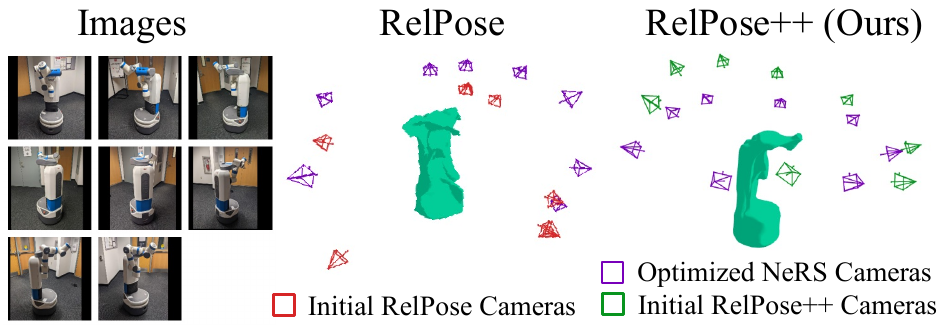}
    \caption{\textbf{Sparse-view 3D Reconstruction using NeRS.} We find that the camera poses estimated by our method are sufficient as initialization for 3D reconstruction. We compare our recovered cameras (green) with RelPose cameras (red) as initialization to NeRS. NeRS jointly optimizes these cameras and shape. We visualize the cameras at the end of the NeRS optimization in purple. We find that our cameras enable higher-fidelity 3D reconstruction.}
    \label{fig:ners}
\end{figure}

\begin{table}
\begin{center}
\footnotesize

\begin{tabular}{cccccccc}
    \toprule
    & & \multicolumn{3}{c}{Cam. Cen.} & \multicolumn{3}{c}{Transl.} \\
    \cmidrule(lr){3-5} \cmidrule(lr){6-8}
    & \# of Images & 3 & 5 & 8 & 3 & 5 & 8\\
    \midrule
    \parbox[t]{2mm}{\multirow{2}{*}{\rotatebox[origin=c]{90}{Seen}}} & Ours & 92.3 & 87.5 & 85.5 & 90.5 & 87.8 & 86.2\\
    & Constant & 91.3 & 84.7 & 81.1 & 69.0 & 60.8 & 56.5\\
    \midrule
    \parbox[t]{2mm}{\multirow{2}{*}{\rotatebox[origin=c]{90}{Uns.}}} & Ours & 82.5 & 71.9 & 67.5 & 79.7 & 74.9 & 73.6\\
    & Constant & 81.4 & 69.5 & 63.8 & 60.3 & 52.2 & 48.2 \\
    \bottomrule%
\end{tabular}%
\end{center}
\caption{\textbf{Analyzing Translation Prediction.}
We quantify the improvements of our predicted translations over a naive baseline that predicts center-facing cameras located at a unit distance from the origin. Because the camera center entangles the rotation and translation prediction, we compute an additional translation accuracy that reports the fraction of translations within 0.1 of the scene scale of the ground truth translation after applying a scaling and world origin alignment (see supplement).
}
\label{tab:translation}

\end{table}

\begin{table}[]
    \centering
    \footnotesize
    \begin{tabular}{ccccccc}
        \toprule
        & \multicolumn{3}{c}{Rotation} & \multicolumn{3}{c}{Cam. Cen.} \\
        \cmidrule(lr){2-4} \cmidrule(lr){5-7}
        \# of Images & 3 & 5 & 8 & 3 & 5 & 8\\
        \midrule
        MediaPipe~\cite{Lugaresi2019MediaPipeAF} & 52.3 & 52.8 & 52.7 & 74.5 & 59.1 & 49.9 \\
        PoseDiffusion~\cite{wang2023posediffusion} & 69.2 & 68.0 & 70.0 & 87.2 & 73.8 & 67.2 \\
        Ours & \textbf{75.8} & \textbf{76.6} & \textbf{77.0} & \textbf{91.6} & \textbf{83.9} & \textbf{77.6}\\
        \bottomrule \\
    \end{tabular}
    \caption{\textbf{Evaluating Zero-shot Generalization on Objectron on Rotation (@ 15\degree) and Camera Center (@ 0.2) Accuracy.} We evaluate our approach, trained on CO3D, on Objectron without any fine-tuning.}
    \label{tab:objectron}
\end{table}

\parhead{Variants.}
We also report comparisons to variants of our approach to highlight the benefits of the energy-based prediction, multi-view reasoning, and proposed translation coordinate frame.

\parsub{Pose Regression.} This corresponds to a regression approach that uses our ResNet and Transformer architecture to directly predict the global rotations (assuming the first camera has identity rotation) and translations. This rotation prediction is analogous to the initial regressor in SparsePose~\cite{sinha23sparsepose} (unfortunately, due to licensing issues, we were unable to obtain code/models for direct comparison). We consider variants that regress translations using the first camera frame and our look-at-centered coordinate frame.

\parsub{Ours (N=2).} This represents a variant that only has access to 2 images at a time when inferring pairwise rotation distributions. While similar to RelPose, it helps disambiguate the benefits of our transformer-based architecture.

\parsub{First-frame Centered Regression.} Instead of using the Look-at centered world frame, this variant defines camera translations using the first camera as world origin (while using the same scaling as the Look-at centered system).

\subsection{Quantitative Results}
\vspace{-2mm}

\parhead{Accuracy of Recovered 6D Poses.}
We evaluate rotation accuracy in \cref{tab:joint_rotations}. We find that our approach significantly outperforms COLMAP and RelPose. While COLMAP performs well at fine error thresholds (see the supplement), it frequently does not converge in sparse-view settings because wide baselines do not provide enough overlap to compute useful correspondences. We find that the Pose Regression baseline performs poorly, suggesting that modeling uncertainty is important for sparse-view settings. The jump in performance from RelPose to our $N=2$ variant suggests that the increased capacity of our transformer architecture is important. Finally, we find that our model starts out at a similar performance to the $N=2$ model but quickly outperforms it for larger $N$, suggesting that the image context is important. Our method also consistently achieves better localization than PoseDiffusion.

We evaluate the camera center accuracy in \cref{tab:camera_center} and also report AUC metrics in the supplement. COLMAP performs poorly since it often fails to converge. 
We find that the First Camera Frame Regression has the worst generalization to unseen object categories (see \cref{tab:translation}). This makes sense because the predicted translation must also account for any errors in the predicted rotation, which likely occur in a different distribution than seen for training categories.

\parhead{Analyzing Translation Predictions.} 
While the focus of our work is on roughly center-facing images of objects as these captures most closely resemble a typical object scanning pipeline, we do find that CO3D deviates significantly from perfectly circular trajectories. We quantify this using an additional baseline that uses our rotation predictions but always predicts a constant [0, 0, 1] translation (which would be optimal for center-facing cameras on a sphere). In addition to camera center evaluation which conflates the predicted rotation and translation, we propose a translation evaluation that computes the proportion of predicted translations that are within 10\% of the ground truth translation. Similar to the camera center evaluation, we apply an optimal similarity transform that accounts for the scene scaling and world origin placement (see supplement for more details). We find that our method significantly outperforms the constant translation baseline using both the camera center and translation metrics in \cref{tab:translation} (reducing translation error from 51.8\% to 26.4\%).

\parhead{Evaluating Generalization.} We evaluate zero-shot generalization on Objectron~\cite{ahmadyan2021objectron} in \cref{tab:objectron}, and find that our approach outperforms PoseDiffusion~\cite{wang2023posediffusion}. We also report the accuracy of relative poses recovered from a per-frame 6D pose estimation method Media\-Pipe~\cite{Lugaresi2019MediaPipeAF}. Note that both our model and PoseDiffusion are trained only on CO3D with no finetuning while MediaPipe is trained per category on Objectron.
Following PoseDiffusion, we also evaluate generalization via zero-shot transfer to RealEstate10K~\cite{zhou2018stereo} and outperform their method, but this scene-level front-facing dataset is not an ideal testbed for testing generalization from 360-degree object-centric data as even a naive baseline (fixed identity rotation) performs well (see supplement).

\subsection{Qualitative Results}
\vspace{-2mm}

\parhead{Visualization for Co3D Predictions.} We compared recovered camera poses from sparse-view images using our method with COLMAP (with SuperPoint/SuperGlue) and RelPose in \cref{fig:qualitative}. We find that our method is able to recover more accurate cameras than RelPose consistently. While COLMAP recovers highly accurate trajectories when it succeeds, it usually fails to converge for sparse images.

We also visualize the effect of increasing image context on pairwise rotation distributions in \cref{fig:resolve_ambiguity}. Given just two images, the relative pose is often ambiguous, but we find that this ambiguity can be resolved by conditioning on more images using our transformer. 

\parhead{In-the-wild Generalization and 3D Reconstruction.} 
We demonstrate the generalization of RelPose++ on in-the-wild captures in \cref{fig:unconstrained_objects}. These  recovered cameras are sufficient to enable 3D reconstruction using NeRS~\cite{zhang2021ners}, a representative sparse-view reconstruction method (\cref{fig:ners}).
\section{Discussion}
We presented \ours{}, a system for inferring a consistent set of 6D poses (rotations and translations) given a sparse set of input views.
While it can robustly infer camera poses, these are not as precise as ones obtained from classical methods, and can be improved further via refinement~\cite{sinha23sparsepose,lin2021barf}. Secondly, while the energy-based models can efficiently capture uncertainty, they are inefficient to sample from and are limited to pairwise distributions, and it may be possible to instead leverage diffusion models to overcome these limitations. Lastly, while we demonstrated that our estimated poses can enable downstream 3D reconstruction, it would be beneficial to develop unified approaches that jointly tackle the tasks of reconstruction and pose inference.

\parhead{Acknowledgements:} We would like to thank Samarth Sinha for useful discussion and thank Sudeep Dasari and Helen Jiang for their feedback on drafts of the paper. This work was supported in part by the NSF GFRP (Grant No. DGE1745016), a CISCO gift award, and the Intelligence Advanced Research Projects Activity (IARPA) via Department of Interior/Interior Business Center (DOI/IBC) contract number 140D0423C0074. The U.S. Government is authorized to reproduce and distribute reprints for Governmental purposes notwithstanding any copyright annotation thereon. 
Disclaimer: The views and conclusions contained herein are those of the authors and should not be interpreted as necessarily representing the official policies or endorsements, either expressed or implied, of IARPA, DOI/IBC, or the U.S. Government.

{
    \small
    \bibliographystyle{ieeenat_fullname}
    \bibliography{main}

\begin{thebibliography}{65}
\providecommand{\natexlab}[1]{#1}
\providecommand{\url}[1]{\texttt{#1}}
\expandafter\ifx\csname urlstyle\endcsname\relax
  \providecommand{\doi}[1]{doi: #1}\else
  \providecommand{\doi}{doi: \begingroup \urlstyle{rm}\Url}\fi

\bibitem[Ahmadyan et~al.(2021)Ahmadyan, Zhang, Ablavatski, Wei, and
  Grundmann]{ahmadyan2021objectron}
Adel Ahmadyan, Liangkai Zhang, Artsiom Ablavatski, Jianing Wei, and Matthias
  Grundmann.
\newblock {Objectron: A Large Scale Dataset of Object-Centric Videos in the
  Wild with Pose Annotations}.
\newblock In \emph{CVPR}, 2021.

\bibitem[Angtian et~al.(2021)Angtian, Kortylewski, and Yuille]{wang2020NeMo}
Wang Angtian, Adam Kortylewski, and Alan Yuille.
\newblock {NeMo: Neural Mesh Models of Contrastive Features for Robust 3D Pose
  Estimation}.
\newblock In \emph{ICLR}, 2021.

\bibitem[Bay et~al.(2006)Bay, Tuytelaars, and Gool]{bay2006surf}
Herbert Bay, Tinne Tuytelaars, and Luc~Van Gool.
\newblock {SURF: Speeded Up Robust Features}.
\newblock In \emph{ECCV}, 2006.

\bibitem[Cai et~al.(2021)Cai, Hariharan, Snavely, and
  Averbuch-Elor]{cai2021extreme}
Ruojin Cai, Bharath Hariharan, Noah Snavely, and Hadar Averbuch-Elor.
\newblock {Extreme Rotation Estimation using Dense Correlation Volumes}.
\newblock In \emph{CVPR}, 2021.

\bibitem[Campos et~al.(2021)Campos, Elvira, G\'omez, Montiel, and
  Tard\'os]{ORBSLAM3_TRO}
Carlos Campos, Richard Elvira, Juan~J. G\'omez, Jos\'e M.~M. Montiel, and
  Juan~D. Tard\'os.
\newblock {ORB-SLAM3: An Accurate Open-Source Library for Visual,
  Visual-Inertial and Multi-Map SLAM}.
\newblock \emph{T-RO}, 2021.

\bibitem[Chang et~al.(2015)Chang, Funkhouser, Guibas, Hanrahan, Huang, Li,
  Savarese, Savva, Song, Su, et~al.]{chang2015shapenet}
Angel~X Chang, Thomas Funkhouser, Leonidas Guibas, Pat Hanrahan, Qixing Huang,
  Zimo Li, Silvio Savarese, Manolis Savva, Shuran Song, Hao Su, et~al.
\newblock {ShapeNet: An Information-Rich 3D Model Repository}.
\newblock \emph{arXiv preprint arXiv:1512.03012}, 2015.

\bibitem[Chen et~al.(2021)Chen, Snavely, and Makadia]{chen2021wide}
Kefan Chen, Noah Snavely, and Ameesh Makadia.
\newblock {Wide-Baseline Relative Camera Pose Estimation with Directional
  Learning}.
\newblock In \emph{CVPR}, 2021.

\bibitem[Choy et~al.(2016)Choy, Gwak, Savarese, and
  Chandraker]{choy2016universal}
Christopher~B Choy, JunYoung Gwak, Silvio Savarese, and Manmohan Chandraker.
\newblock {Universal Correspondence Network}.
\newblock \emph{NeurIPS}, 2016.

\bibitem[Davison et~al.(2007)Davison, Reid, Molton, and
  Stasse]{davison2007monoslam}
Andrew~J Davison, Ian~D Reid, Nicholas~D Molton, and Olivier Stasse.
\newblock {MonoSLAM: Real-time Single Camera SLAM}.
\newblock \emph{TPAMI}, 2007.

\bibitem[DeTone et~al.(2018)DeTone, Malisiewicz, and
  Rabinovich]{detone2018superpoint}
Daniel DeTone, Tomasz Malisiewicz, and Andrew Rabinovich.
\newblock {SuperPoint: Self-supervised Interest Point Detection and
  Description}.
\newblock In \emph{CVPR-W}, 2018.

\bibitem[Fischler and Bolles(1981)]{fischler1981random}
Martin~A Fischler and Robert~C Bolles.
\newblock {Random Sample Consensus: A Paradigm for Model Fitting with
  Applications to Image Analysis and Automated Cartography}.
\newblock \emph{Communications of the ACM}, 1981.

\bibitem[Gilitschenski et~al.(2019)Gilitschenski, Sahoo, Schwarting, Amini,
  Karaman, and Rus]{gilitschenski2019deep}
Igor Gilitschenski, Roshni Sahoo, Wilko Schwarting, Alexander Amini, Sertac
  Karaman, and Daniela Rus.
\newblock {Deep Orientation Uncertainty Learning Based on a Bingham Loss}.
\newblock In \emph{ICLR}, 2019.

\bibitem[Hartley and Zisserman(2003)]{hartley2003multiple}
Richard Hartley and Andrew Zisserman.
\newblock \emph{{Multiple View Geometry in Computer Vision}}.
\newblock {Cambridge University Press}, 2003.

\bibitem[Hartley et~al.(2013)Hartley, Trumpf, Dai, and Li]{hartley2013rotation}
Richard Hartley, Jochen Trumpf, Yuchao Dai, and Hongdong Li.
\newblock {Rotation Averaging}.
\newblock \emph{IJCV}, 2013.

\bibitem[He et~al.(2016)He, Zhang, Ren, and Sun]{he2016deep}
Kaiming He, Xiangyu Zhang, Shaoqing Ren, and Jian Sun.
\newblock {Deep Residual Learning for Image Recognition}.
\newblock In \emph{CVPR}, 2016.

\bibitem[Jiang et~al.(2022)Jiang, Jiang, Grauman, and Zhu]{jiang2022forge}
Hanwen Jiang, Zhenyu Jiang, Kristen Grauman, and Yuke Zhu.
\newblock {Few-View Object Reconstruction with Unknown Categories and Camera
  Poses}.
\newblock \emph{ArXiv}, 2212.04492, 2022.

\bibitem[Kanazawa et~al.(2018)Kanazawa, Black, Jacobs, and
  Malik]{KanazawaHMR17}
Angjoo Kanazawa, Michael~J. Black, David~W. Jacobs, and Jitendra Malik.
\newblock {End-to-end Recovery of Human Shape and Pose}.
\newblock In \emph{CVPR}, 2018.

\bibitem[Kanazawa et~al.(2019)Kanazawa, Zhang, Felsen, and
  Malik]{kanazawa2019learning}
Angjoo Kanazawa, Jason~Y Zhang, Panna Felsen, and Jitendra Malik.
\newblock {Learning 3D Human Dynamics from Video}.
\newblock In \emph{CVPR}, 2019.

\bibitem[Kocabas et~al.(2020)Kocabas, Athanasiou, and Black]{kocabas2020vibe}
Muhammed Kocabas, Nikos Athanasiou, and Michael~J Black.
\newblock {VIBE: Video Inference for Human Body Pose and Shape Estimation}.
\newblock In \emph{CVPR}, 2020.

\bibitem[Lin et~al.(2021)Lin, Ma, Torralba, and Lucey]{lin2021barf}
Chen-Hsuan Lin, Wei-Chiu Ma, Antonio Torralba, and Simon Lucey.
\newblock {BARF: Bundle-Adjusting Neural Radiance Fields}.
\newblock In \emph{ICCV}, 2021.

\bibitem[Liu et~al.(2010)Liu, Yuen, and Torralba]{liu2010sift}
Ce Liu, Jenny Yuen, and Antonio Torralba.
\newblock {SIFT Flow: Dense Correspondence Across Scenes and Its Applications}.
\newblock \emph{TPAMI}, 2010.

\bibitem[Longuet-Higgins(1981)]{longuet1981computer}
H~Christopher Longuet-Higgins.
\newblock {A Computer Algorithm for Reconstructing a Scene from Two
  Projections}.
\newblock \emph{Nature}, 1981.

\bibitem[Lowe(2004)]{lowe2004distinctive}
David~G Lowe.
\newblock {Distinctive Image Features from Scale-invariant Keypoints}.
\newblock \emph{IJCV}, 2004.

\bibitem[Lucas and Kanade(1981)]{lucas1981iterative}
Bruce~D Lucas and Takeo Kanade.
\newblock {An Iterative Image Registration Technique with an Application to
  Stereo Vision}.
\newblock In \emph{IJCAI}, 1981.

\bibitem[Lugaresi et~al.(2019)Lugaresi, Tang, Nash, McClanahan, Uboweja, Hays,
  Zhang, Chang, Yong, Lee, Chang, Hua, Georg, and
  Grundmann]{Lugaresi2019MediaPipeAF}
Camillo Lugaresi, Jiuqiang Tang, Hadon Nash, Chris McClanahan, Esha Uboweja,
  Michael Hays, Fan Zhang, Chuo-Ling Chang, Ming~Guang Yong, Juhyun Lee,
  Wan-Teh Chang, Wei Hua, Manfred Georg, and Matthias Grundmann.
\newblock {MediaPipe: A Framework for Building Perception Pipelines}.
\newblock \emph{arXiv:1906.08172}, 2019.

\bibitem[Ma et~al.(2022)Ma, Yang, Wang, Urtasun, and Torralba]{ma2022virtual}
Wei-Chiu Ma, Anqi~Joyce Yang, Shenlong Wang, Raquel Urtasun, and Antonio
  Torralba.
\newblock {Virtual Correspondence: Humans as a Cue for Extreme-View Geometry}.
\newblock In \emph{CVPR}, 2022.

\bibitem[Manhardt et~al.(2019)Manhardt, Arroyo, Rupprecht, Busam, Birdal,
  Navab, and Tombari]{manhardt2019explaining}
Fabian Manhardt, Diego~Martin Arroyo, Christian Rupprecht, Benjamin Busam,
  Tolga Birdal, Nassir Navab, and Federico Tombari.
\newblock {Explaining the Ambiguity of Object Detection and 6D Pose from Visual
  Data}.
\newblock In \emph{ICCV}, 2019.

\bibitem[Mehta et~al.(2017)Mehta, Sridhar, Sotnychenko, Rhodin, Shafiei,
  Seidel, Xu, Casas, and Theobalt]{mehta2017vnect}
Dushyant Mehta, Srinath Sridhar, Oleksandr Sotnychenko, Helge Rhodin, Mohammad
  Shafiei, Hans-Peter Seidel, Weipeng Xu, Dan Casas, and Christian Theobalt.
\newblock {VNect: Real-time 3D Human Pose Estimation with a Single RGB Camera}.
\newblock \emph{TOG}, 2017.

\bibitem[Melekhov et~al.(2017)Melekhov, Ylioinas, Kannala, and
  Rahtu]{melekhov2017relative}
Iaroslav Melekhov, Juha Ylioinas, Juho Kannala, and Esa Rahtu.
\newblock {Relative Camera Pose Estimation Using Convolutional Neural
  Networks}.
\newblock In \emph{ACIVS}, 2017.

\bibitem[Mildenhall et~al.(2020)Mildenhall, Srinivasan, Tancik, Barron,
  Ramamoorthi, and Ng]{mildenhall2020nerf}
Ben Mildenhall, Pratul~P. Srinivasan, Matthew Tancik, Jonathan~T. Barron, Ravi
  Ramamoorthi, and Ren Ng.
\newblock {NeRF: Representing Scenes as Neural Radiance Fields for View
  Synthesis}.
\newblock In \emph{ECCV}, 2020.

\bibitem[Mur-Artal and Tard\'os(2017)]{murORB2}
Ra\'ul Mur-Artal and Juan~D. Tard\'os.
\newblock {ORB-SLAM2: An Open-Source {SLAM} System for Monocular, Stereo and
  {RGB-D} Cameras}.
\newblock \emph{T-RO}, 2017.

\bibitem[Mur-Artal et~al.(2015)Mur-Artal, Montiel, and Tardos]{mur2015orb}
Raul Mur-Artal, Jose Maria~Martinez Montiel, and Juan~D Tardos.
\newblock {ORB-SLAM: A Versatile and Accurate Monocular SLAM System}.
\newblock \emph{T-RO}, 2015.

\bibitem[Murphy et~al.(2021)Murphy, Esteves, Jampani, Ramalingam, and
  Makadia]{implicitpdf2021}
Kieran~A Murphy, Carlos Esteves, Varun Jampani, Srikumar Ramalingam, and Ameesh
  Makadia.
\newblock {Implicit-PDF: Non-Parametric Representation of Probability
  Distributions on the Rotation Manifold}.
\newblock In \emph{ICML}, 2021.

\bibitem[Nguyen et~al.(2022)Nguyen, Du, Xiao, Ramamonjisoa, and
  Lepetit]{nguyen2022pizza}
Van~Nguyen Nguyen, Yuming Du, Yang Xiao, Michael Ramamonjisoa, and Vincent
  Lepetit.
\newblock {PIZZA: A Powerful Image-only Zero-Shot Zero-CAD Approach to 6 DoF
  Tracking}.
\newblock In \emph{3DV}, 2022.

\bibitem[Nist{\'e}r(2004)]{nister2004efficient}
David Nist{\'e}r.
\newblock {An Efficient Solution to the Five-point Relative Pose Problem}.
\newblock \emph{TPAMI}, 2004.

\bibitem[Novotny et~al.(2017)Novotny, Larlus, and Vedaldi]{novotny2017learning}
David Novotny, Diane Larlus, and Andrea Vedaldi.
\newblock {Learning 3D Object Categories by Looking Around Them}.
\newblock In \emph{ICCV}, 2017.

\bibitem[Okorn et~al.(2020)Okorn, Xu, Hebert, and Held]{okorn2020learning}
Brian Okorn, Mengyun Xu, Martial Hebert, and David Held.
\newblock {Learning Orientation Distributions for Object Pose Estimation}.
\newblock In \emph{IROS}, 2020.

\bibitem[Okorn et~al.(2021)Okorn, Gu, Hebert, and Held]{okorn2021zephyr}
Brian Okorn, Qiao Gu, Martial Hebert, and David Held.
\newblock {ZePHyR: Zero-shot Pose Hypothesis Scoring}.
\newblock In \emph{ICRA}, 2021.

\bibitem[Reizenstein et~al.(2021)Reizenstein, Shapovalov, Henzler, Sbordone,
  Labatut, and Novotny]{reizenstein21co3d}
Jeremy Reizenstein, Roman Shapovalov, Philipp Henzler, Luca Sbordone, Patrick
  Labatut, and David Novotny.
\newblock {Common Objects in 3D: Large-Scale Learning and Evaluation of
  Real-life 3D Category Reconstruction}.
\newblock In \emph{ICCV}, 2021.

\bibitem[Rockwell et~al.(2022)Rockwell, Johnson, and Fouhey]{rockwell20228}
Chris Rockwell, Justin Johnson, and David~F Fouhey.
\newblock {The 8-Point Algorithm as an Inductive Bias for Relative Pose
  Prediction by ViTs}.
\newblock In \emph{{3DV}}, 2022.

\bibitem[Sarlin et~al.(2019)Sarlin, Cadena, Siegwart, and
  Dymczyk]{sarlin2019coarse}
Paul-Edouard Sarlin, Cesar Cadena, Roland Siegwart, and Marcin Dymczyk.
\newblock {From Coarse to Fine: Robust Hierarchical Localization at Large
  Scale}.
\newblock In \emph{CVPR}, 2019.

\bibitem[Sarlin et~al.(2020)Sarlin, DeTone, Malisiewicz, and
  Rabinovich]{sarlin2020superglue}
Paul-Edouard Sarlin, Daniel DeTone, Tomasz Malisiewicz, and Andrew Rabinovich.
\newblock {SuperGlue: Learning Feature Matching with Graph Neural Networks}.
\newblock In \emph{CVPR}, 2020.

\bibitem[Sch\"{o}nberger and Frahm(2016)]{schoenberger2016sfm}
Johannes~Lutz Sch\"{o}nberger and Jan-Michael Frahm.
\newblock {Structure-from-Motion Revisited}.
\newblock In \emph{CVPR}, 2016.

\bibitem[Sch\"{o}nberger et~al.(2016)Sch\"{o}nberger, Zheng, Pollefeys, and
  Frahm]{schoenberger2016mvs}
Johannes~Lutz Sch\"{o}nberger, Enliang Zheng, Marc Pollefeys, and Jan-Michael
  Frahm.
\newblock {Pixelwise View Selection for Unstructured Multi-View Stereo}.
\newblock In \emph{ECCV}, 2016.

\bibitem[Sinha et~al.(2023)Sinha, Zhang, Tagliasacchi, Gilitschenski, and
  Lindell]{sinha23sparsepose}
Samarth Sinha, Jason~Y Zhang, Andrea Tagliasacchi, Igor Gilitschenski, and
  David~B Lindell.
\newblock {SparsePose: Sparse-View Camera Pose Regression and Refinement}.
\newblock In \emph{CVPR}, 2023.

\bibitem[Sturm et~al.(2012)Sturm, Engelhard, Endres, Burgard, and
  Cremers]{sturm2012benchmark}
Jürgen Sturm, Nikolas Engelhard, Felix Endres, Wolfram Burgard, and Daniel
  Cremers.
\newblock {A Benchmark for the Evaluation of RGB-D SLAM Systems}.
\newblock In \emph{{IROS}}, 2012.

\bibitem[Sun et~al.(2021)Sun, Tagliasacchi, Deng, Sabour, Yazdani, Hinton, and
  Yi]{sun2021canonical}
Weiwei Sun, Andrea Tagliasacchi, Boyang Deng, Sara Sabour, Soroosh Yazdani,
  Geoffrey~E Hinton, and Kwang~Moo Yi.
\newblock {Canonical Capsules: Self-supervised Capsules in Canonical Pose}.
\newblock In \emph{NeurIPS}, 2021.

\bibitem[Teed and Deng(2021)]{teed2021droid}
Zachary Teed and Jia Deng.
\newblock {DROID-SLAM: Deep Visual SLAM for Monocular, Stereo, and RGB-D
  Cameras}.
\newblock \emph{NeurIPS}, 2021.

\bibitem[Triggs et~al.(1999)Triggs, McLauchlan, Hartley, and
  Fitzgibbon]{triggs1999bundle}
Bill Triggs, Philip~F McLauchlan, Richard~I Hartley, and Andrew~W Fitzgibbon.
\newblock {Bundle Adjustment---A Modern Synthesis}.
\newblock In \emph{International workshop on vision algorithms}, 1999.

\bibitem[Umeyama(1991)]{umeyama1991least}
Shinji Umeyama.
\newblock {Least-squares Estimation of Transformation Parameters Between Two
  Point Patterns}.
\newblock \emph{TPAMI}, 1991.

\bibitem[Usman et~al.(2022)Usman, Tagliasacchi, Saenko, and
  Sud]{usman2022metapose}
Ben Usman, Andrea Tagliasacchi, Kate Saenko, and Avneesh Sud.
\newblock {MetaPose: Fast 3D Pose from Multiple Views without 3D Supervision}.
\newblock In \emph{CVPR}, 2022.

\bibitem[Vaswani et~al.(2017)Vaswani, Shazeer, Parmar, Uszkoreit, Jones, Gomez,
  Kaiser, and Polosukhin]{vaswani2017attention}
Ashish Vaswani, Noam Shazeer, Niki Parmar, Jakob Uszkoreit, Llion Jones,
  Aidan~N Gomez, {\L}ukasz Kaiser, and Illia Polosukhin.
\newblock {Attention is All You Need}.
\newblock \emph{NeurIPS}, 2017.

\bibitem[Wang et~al.(2023)Wang, Rupprecht, and Novotny]{wang2023posediffusion}
Jianyuan Wang, Christian Rupprecht, and David Novotny.
\newblock {PoseDiffusion: Solving Pose Estimation via Diffusion-aided Bundle
  Adjustment}.
\newblock In \emph{ICCV}, 2023.

\bibitem[Wang et~al.(2017)Wang, Clark, Wen, and Trigoni]{wang2017deepvo}
Sen Wang, Ronald Clark, Hongkai Wen, and Niki Trigoni.
\newblock {DeepVO: Towards End-to-End Visual Odometry with Deep Recurrent
  Convolutional Neural Networks}.
\newblock In \emph{ICRA}, 2017.

\bibitem[Wong et~al.(2017)Wong, Kee, Le, Wagner, Mariottini, Schneider,
  Hamilton, Chipalkatty, Hebert, Johnson, et~al.]{wong2017segicp}
Jay~M Wong, Vincent Kee, Tiffany Le, Syler Wagner, Gian-Luca Mariottini,
  Abraham Schneider, Lei Hamilton, Rahul Chipalkatty, Mitchell Hebert, David~MS
  Johnson, et~al.
\newblock {SegICP: Integrated Deep Semantic Segmentation and Pose Estimation}.
\newblock In \emph{IROS}, 2017.

\bibitem[Xiao et~al.(2019)Xiao, Qiu, Langlois, Aubry, and
  Marlet]{Xiao2019PoseFromShape}
Yang Xiao, Xuchong Qiu, Pierre{-}Alain Langlois, Mathieu Aubry, and Renaud
  Marlet.
\newblock {Pose from Shape: Deep Pose Estimation for Arbitrary {3D} Objects}.
\newblock In \emph{BMVC}, 2019.

\bibitem[Xiao et~al.(2021)Xiao, Du, and Marlet]{Xiao2020PoseContrast}
Yang Xiao, Yuming Du, and Renaud Marlet.
\newblock {PoseContrast: Class-Agnostic Object Viewpoint Estimation in the Wild
  with Pose-Aware Contrastive Learning}.
\newblock In \emph{3DV}, 2021.

\bibitem[Yang et~al.(2020)Yang, Stumberg, Wang, and Cremers]{yang2020d3vo}
Nan Yang, Lukas~von Stumberg, Rui Wang, and Daniel Cremers.
\newblock {D3VO: Deep Depth, Deep Pose and Deep Uncertainty for Monocular
  Visual Odometry}.
\newblock In \emph{CVPR}, 2020.

\bibitem[Yu et~al.(2021)Yu, Ye, Tancik, and Kanazawa]{yu2021pixelnerf}
Alex Yu, Vickie Ye, Matthew Tancik, and Angjoo Kanazawa.
\newblock {pixelNeRF: Neural Radiance Fields from One or Few Images}.
\newblock In \emph{CVPR}, 2021.

\bibitem[Zhang et~al.(2020)Zhang, Pepose, Joo, Ramanan, Malik, and
  Kanazawa]{zhang2020phosa}
Jason~Y. Zhang, Sam Pepose, Hanbyul Joo, Deva Ramanan, Jitendra Malik, and
  Angjoo Kanazawa.
\newblock {Perceiving 3D Human-Object Spatial Arrangements from a Single Image
  in the Wild}.
\newblock In \emph{ECCV}, 2020.

\bibitem[Zhang et~al.(2021)Zhang, Yang, Tulsiani, and Ramanan]{zhang2021ners}
Jason~Y. Zhang, Gengshan Yang, Shubham Tulsiani, and Deva Ramanan.
\newblock {NeRS: Neural Reflectance Surfaces for Sparse-view 3D Reconstruction
  in the Wild}.
\newblock In \emph{NeurIPS}, 2021.

\bibitem[Zhang et~al.(2022)Zhang, Ramanan, and Tulsiani]{zhang2022relpose}
Jason~Y. Zhang, Deva Ramanan, and Shubham Tulsiani.
\newblock {RelPose: Predicting Probabilistic Relative Rotation for Single
  Objects in the Wild}.
\newblock In \emph{ECCV}, 2022.

\bibitem[Zhang(2019)]{zhang2019shiftinvar}
Richard Zhang.
\newblock {Making Convolutional Networks Shift-Invariant Again}.
\newblock In \emph{ICML}, 2019.

\bibitem[Zhou et~al.(2018)Zhou, Tucker, Flynn, Fyffe, and
  Snavely]{zhou2018stereo}
Tinghui Zhou, Richard Tucker, John Flynn, Graham Fyffe, and Noah Snavely.
\newblock Stereo magnification: Learning view synthesis using multiplane
  images.
\newblock \emph{SIGGRAPH}, 37, 2018.

\bibitem[Zhou and Tulsiani(2023)]{zhou2022sparsefusion}
Zhizhuo Zhou and Shubham Tulsiani.
\newblock {SparseFusion: Distilling View-conditioned Diffusion for 3D
  Reconstruction}.
\newblock In \emph{CVPR}, 2023.

\end{thebibliography}
}
\clearpage
\setcounter{page}{1}
\maketitlesupplementary

\section{Appendix}

\subsection{Generalization Experiments, Ablations, and Evaluations}

\parhead{Generalization to RealEstate10K dataset.} 
Following PoseDiffusion~\cite{wang2023posediffusion}, we evaluate the zero-shot generalization of our method trained only on CO3D in \cref{tab:realestate10k}. Similar to \cite{wang2023posediffusion}, we sample 2000 sequences to reduce evaluation time (note that the sampled sequences differ, which may explain the difference in the reported results). We find that evaluation on RealEstate10K is not particularly informative as the trajectories tend to be largely front-facing. As a result, blindly predicting identity rotation (Constant Rot.) performs well, even surpassing PoseDiffusion in performance. 

\parhead{Positional Encoding Ablation.} Our transformer encoder takes the positionally-encoded image index as input in addition to the image features. Although our model does not expect the images to be ordered (at training time, the order is randomized), we find that the image index positional encoding is helpful in \cref{tab:positional_encoding}. We hypothesize that this helps disambiguate images, particularly the first one.

\begin{figure}
    \centering
    \includegraphics[width=\linewidth]{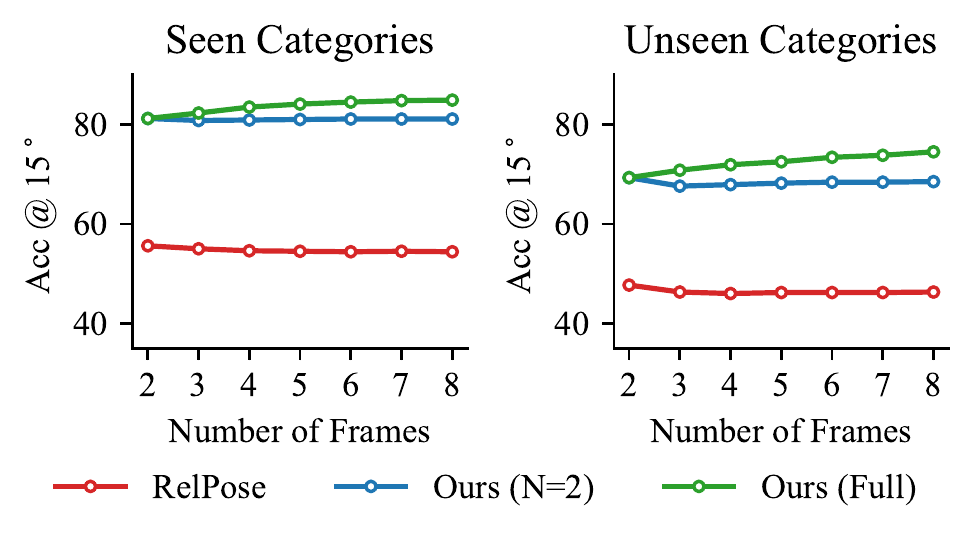}
    \caption{\textbf{Pairwise Rotation Accuracy.} We report pairwise rotation accuracy as the percent of predicted relative rotations within 15 degrees of the true relative rotation for every pair of cameras in the scene. Our (full) model exhibits increased performance with the number of frames. The ability to show our model more than just two images at a time provides valuable scene context that can help with resolving pose ambiguity caused by object symmetry.}
    \label{fig:pairwise_accuracy}
\end{figure}

\parhead{Analyzing Pairwise Rotation Distributions.} In our main rotation and camera center evaluations, we aim to recover $N$ globally consistent cameras given $N$ images. To evaluate the importance of context, we also evaluate the rotation accuracy for pairs of images, possibly conditioned on more than 2 images. We evaluate the proportion of pairwise relative poses within 15 degrees of the ground truth in \cref{fig:pairwise_accuracy}.
We find that our transformer-based architecture can use the context of additional images to improve pairwise accuracy.

\subsection{Implementation Details}

\begin{figure*}[t]
    \centering
    \includegraphics[width=\textwidth]{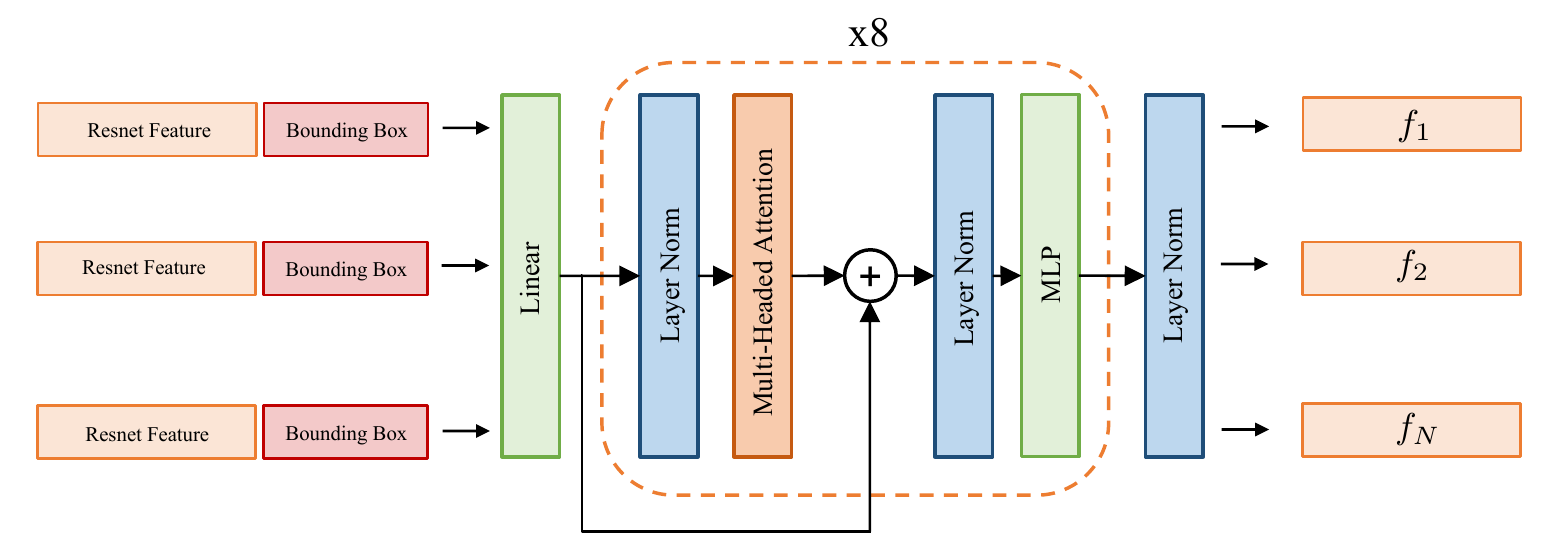}
    \caption{\textbf{Encoder Architecture Diagram} Our transformer encoder takes as input a set of N Resnet image features concatenated with the bounding box information for that image. After 8 layers of 12-headed self-attention and a final Layer Normalization layer, we receive output features $f_1$ ... $f_N$. Unlike the input Resnet features, each of these features $f_1$ ... $f_N$ depends on every image in the set.
}
    \label{fig:encoder}
\end{figure*}

\begin{figure*}[t]
    \centering
    \includegraphics[width=\textwidth]{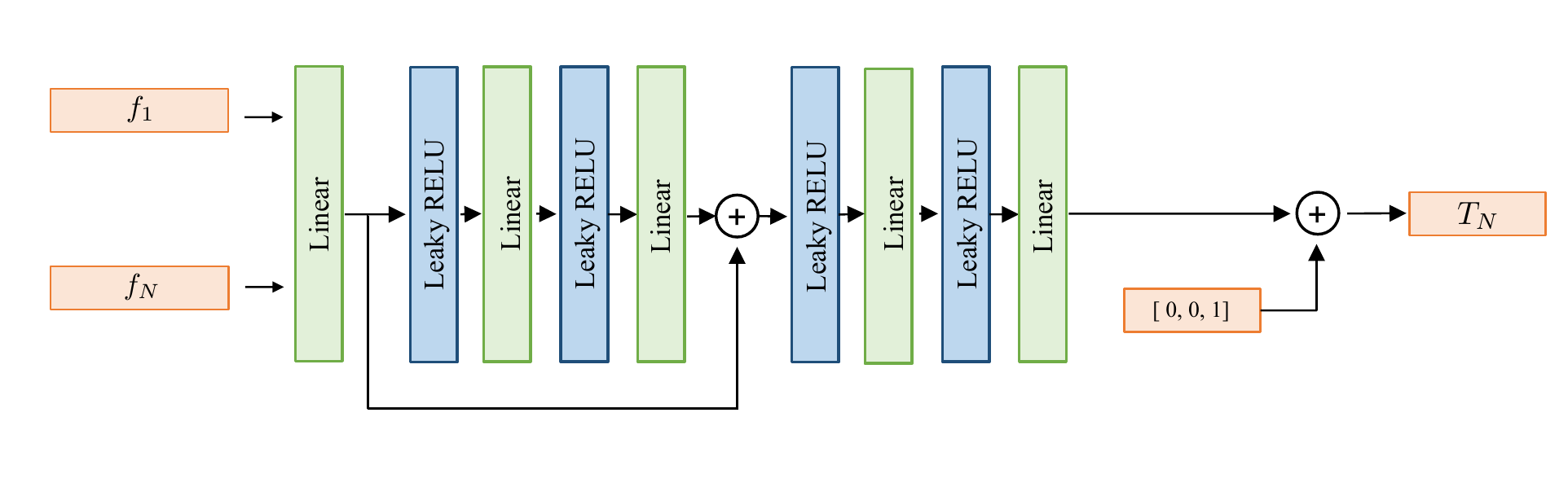}
    \caption{\textbf{Translation Regressor Architecture Diagram} For regressing to $T_N$, we concatenate transformer features $f_1$ and $f_N$ as input to the translation regressor network. We choose to include $f_1$ as input since the scale of the scene is determined by the first image (the norm of the translation target for the first camera is always set to 1).}
    \label{fig:regressor}
\end{figure*}

\paragraph{Network Details.} 
An anti-aliased~\cite{zhang2019shiftinvar} ResNet-50~\cite{he2016deep} is used to extract image features. These image features are then positionally encoded by image index, concatenated with bounding box parameters, and passed as input to our encoder, the architecture of which is shown in \cref{fig:encoder}. The pairwise rotation score predictor $g(f_1, f_2, R)$ is a 4-layer MLP that takes as input concatenated transformer features and outputs a score. The translation regressor architecture is depicted in \cref{fig:regressor}. For the Pose Regression approach, a network similar to the translation regressor is used to predict the 6D rotation representation, instead of the translation residual.

\begin{figure*}
    \centering
    \includegraphics[width=\textwidth]{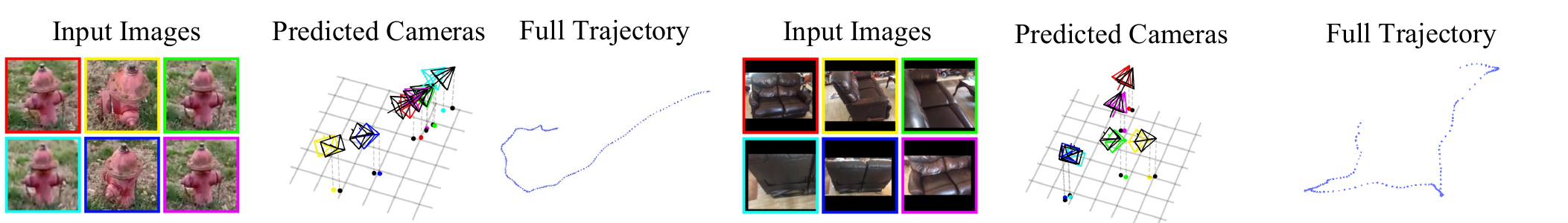}
    \caption{\textbf{Qualitative Results on Non-spherical Camera Trajectories in CO3D.} While CO3D captures tend to be object-centric, the camera trajectories can vary in shape, sometimes requiring challenging translation predictions. Here, we show some predictions on sparse-view sequences as well as the full camera trajectory for reference. }
    \label{fig:noncircular}
\end{figure*}

\paragraph{CO3D Dataset~\cite{reizenstein21co3d}.} Following \cite{zhang2022relpose}, we set aside the following 10 of 51 CO3Dv2 categories to test generalization: ball, book, couch, frisbee, hotdog, kite, remote, sandwich, skateboard, and suitcase. All CO3D evaluations are averaged over 5 runs. For each run, we randomly sample an ordering such that the images for $N$=2 are always a subset of $N$=3 and so on.

\paragraph{RealEstate10K Dataset~\cite{zhou2018stereo}.} We randomly sample 2,000 sequences from the dataset. Our split is likely different than that used by PoseDiffusion~\cite{wang2023posediffusion}.

\paragraph{Objectron Dataset~\cite{ahmadyan2021objectron}.} We evaluate on the test set of the 4 categories (Camera, Chair, Cup, Shoe) for which off-the-shelf pre-trained Objectron models~\cite{Lugaresi2019MediaPipeAF} are available. We skip sequences with more than one object instance. In total, we evaluate on 1081 sequences.

\begin{table}[t]
    \centering
    \footnotesize
    \begin{tabular}{ccccccccc}
        \toprule
        & \# of Images & 2 & 3 & 4 & 5 & 6 & 7 & 8\\ \midrule
         \parbox[t]{2mm}{\multirow{3}{*}{\rotatebox[origin=c]{90}{Rot.}}} & Constant Rot. & \textbf{84.0} & 83.8 & 83.9 & 84.0 & 84.0 & 84.0 & 83.9\\
         & PoseDiffusion & 77.6 & 77.9 & 78.4 & 78.7 & 78.9 & 79.3 & 79.0\\
         & Ours & 83.8 & \textbf{85.1} & \textbf{85.8} & \textbf{86.4} & \textbf{86.5} & \textbf{86.7} & \textbf{86.8} \\
         \midrule
         \parbox[t]{2mm}{\multirow{2}{*}{\rotatebox[origin=c]{90}{CC.}}} 
         & PoseDiffusion & - & \textbf{77.7} & \textbf{65.9} & \textbf{60.1} & \textbf{55.0} & \textbf{52.2} & \textbf{50.2} \\
         & Ours & -  & 71.2 & 60.6 & 54.0 & 49.4 & 47.1 & 45.5 \\
          \bottomrule\\
    \end{tabular}
    \caption{\textbf{Evaluation of Rotation and Camera Center Accuracy on RealEstate10K~\cite{zhou2018stereo}.} The constant rotation baseline always predicts an identity rotation. The dataset has a strong, forward-facing bias.
    }
    \label{tab:realestate10k}
\end{table}

\begin{table}
\centering
\footnotesize
\begin{tabular}{lllccccccc}
\toprule
& & \# of Images & 2 & 3 & 4 & 5 & 6 & 7 & 8\\
\midrule
\parbox[t]{2mm}{\multirow{4}{*}{\rotatebox[origin=c]{90}{Rotation}}} &
\parbox[t]{2mm}{\multirow{2}{*}{\rotatebox[origin=c]{90}{Seen}}}
& Ours (Full)  &   \textbf{81.8} & \textbf{82.8} & \textbf{84.1} & \textbf{84.7} & \textbf{84.9} & \textbf{85.3} & \textbf{85.5}\\
& & Ours (no PE) & 79.7 & 81.6 & 82.6 & 83.4 & 83.8 & 84.2 & 84.3\\ 
\cmidrule(lr){2-10}
& \parbox[t]{2mm}{\multirow{2}{*}{\rotatebox[origin=c]{90}{Un}}}
& Ours (Full) & \textbf{69.8} & \textbf{71.1} & \textbf{71.9} & \textbf{72.8} & \textbf{73.8} & \textbf{74.4} & \textbf{74.9}\\
& & Ours (no PE) & 66.3 & 68.0 & 69.4 & 70.5 & 71.1 & 71.7 & 72.0\\
\midrule
\parbox[t]{2mm}{\multirow{4}{*}{\rotatebox[origin=c]{90}{Cam Center}}} &
\parbox[t]{2mm}{\multirow{2}{*}{\rotatebox[origin=c]{90}{Seen}}}
& Ours (Full)  &   - & \textbf{92.3} & \textbf{89.1} & \textbf{87.5} & \textbf{86.3} & \textbf{85.9} & \textbf{85.5} \\
& & Ours (no PE) &  - & 92.5 & 87.2 & 85.3 & 84.4 & 83.9 & 82.8\\ 
\cmidrule(lr){2-10}
& \parbox[t]{2mm}{\multirow{2}{*}{\rotatebox[origin=c]{90}{Un}}}
& Ours (Full) & - & 82.5 & \textbf{75.6} & \textbf{71.9} & \textbf{69.9} & \textbf{68.5} & \textbf{67.5}\\
& & Ours (no PE) & - & 81.8 & 72.7 & 67.0 & 64.1 & 61.5 & 59.8\\
\bottomrule \\

\end{tabular}
\caption{\textbf{Positional Encoding Ablation on CO3D.} We evaluate our method with and without per-image positional encoding on CO3D in terms of rotation and camera center accuracy. We find that applying positional encoding gives a slight improvement across the board.}
\label{tab:positional_encoding}
\end{table}

\paragraph{Recovering Global Rotations from Pairwise Predictions.}
Our relative rotation predictor $g$ infers the distribution over  pairwise relative rotations but does not directly yield a   globally consistent  set of rotations for all the images within the scene. To obtain these, we follow the optimization procedure proposed in RelPose~\cite{zhang2022relpose}. We briefly summarize this below, and refer the reader to ~\cite{zhang2022relpose} for details.

As this optimization only relies on energies for relative rotations and is agnostic to a canonical global rotation, we can fix the rotation of the first camera as $I$. To obtain an initial set of global rotations, RelPose first constructs a dense graph between images, with each edge representing the most likely relative rotation and the weight of the edge corresponding to the score of this rotation. It then constructs a maximum spanning tree such that only N-1 edges representing relative rotations are preserved, and this spanning tree (with each edge representing a relative rotation) can be used to recover per-image rotations assuming $R_1=I$.
This serves as the initialization for the iterative coordinate ascent search algorithm to optimize the global rotations. We refer the reader to \cite{zhang2022relpose} for more details. Note that, unlike \cite{zhang2022relpose} which predicts the relative rotation distributions by only considering features from image pairs, our approach obtains these features by jointly reasoning over all images.

We also note that the task of optimizing for a globally consistent set of rotations given relative poses is similar in spirit to Rotation Averaging~\cite{hartley2013rotation}. The main difference is that Rotation Averaging assumes a unimodal Gaussian distribution centered at every relative rotation whereas our coordinate-ascent method can accommodate more expressive, often multi-modal probability distributions.

\paragraph{Optimal Placement of World Origin for Translation Prediction.}
We now provide a derivation of why placing the world origin at the intersection of the cameras' optical axes is optimal, and specifically, preferable to placing the origin at the first camera (as prior work does). Our derivation assumes a `look-at' scene, where all of the cameras are `looking at' some point in the world frame $\mathbf{p}^{*}$. More specifically, we assume that while the configuration of the camera pose may be undetermined/ambiguous, it is constrained such that the coordinate of this `look-at' point is fixed w.r.t the camera to $\mathbf{l}_i$ \eg if a camera is moving while looking at a ball 1 unit away, $\mathbf{l}_i = [0,0,1]$.

As before, one can transform a point in world coordinates into camera coordinates with $\mathbf{x}_c^i = R_i \mathbf{x}_w + \mathbf{t}_i$. In the more general case, 
we can then express the `look-at' constraint as a simple linear equation:
\begin{gather}
    \mathbf{l}_i = R_i \mathbf{p}^{*} + \mathbf{t}_i
\end{gather}
If we enforce that the camera-frame coordinate of the look-at point remains fixed under possible transformations \ie $\frac{\partial {\bf  l}_i}{\partial R_i} = 0$, we see that the camera translation varies with rotation and that $||\frac{\partial {\bf t}_i}{\partial R_i}|| \propto ||{\bf p}^{*}||$. This dependence is minimized to be 0 if and only if $\mathbf{p}^*=0$ \ie the `look-at' point is chosen to be the center of the world coordinate system.

\subsection{Evaluation Protocol}

\paragraph{Optimal Alignment for Camera Centers Evaluation.} Given ground truth camera centers $\{\hat{\mathbf{c}}_i \}_{i=1}^N$ and predicted camera centers $\{\mathbf{c}_i\}_{i=1}^N$ where $\mathbf{c}_i, \hat{\mathbf{c}}_i \in \R^3$, we aim to evaluate how close the two are. However, as the world coordinate system (and thus the location of cameras in it) are ambiguous up to a similarity transform, we first compute the optimal similarity transform by solving:
\begin{equation}
    \argmin_{s,R,\mathbf{t}} \sum_{i=1}^N \lVert \hat{\mathbf{c}}_i - (sR\mathbf{c}_i  + \mathbf{t})\rVert^2
\end{equation}
we compute the optimal scale $s\in \R^{+}$, rotation $R\in SO(3)$, and translation $\mathbf{t}\in \R^3$ using the algorithm derived by Umeyama~\cite{umeyama1991least}.

After transforming the predicted camera centers $\Tilde{\mathbf{c}}_i=sR\mathbf{c}_i+\mathbf{t}$, we compute the proportion of camera centers within 0.2 of the scene scale: 
\begin{equation}
    \lVert \hat{\mathbf{c}}_i - \Tilde{\mathbf{c}}_i \rVert < 0.2 \sigma
\end{equation} where $\sigma$ is the distance from the centroid of \textit{all} camera centers in the video sequence to the furthest camera center (as done in \cite{sinha23sparsepose}). Note that this scene scale $\sigma$ is computed with respect to \textit{all} camera poses in the video sequence, whereas the optimal similarity transform is computed with respect to the $N$ cameras used for evaluation.

\paragraph{Optimal Alignment for Camera Translation Evaluation.} We note that the camera center ($\mathbf{c}_i = - {R_i} \mathbf{t}_i$) encompasses both the predicted rotation and translation. We thus also propose to evaluate the predicted camera translation independently by  measuring how close the predictions $\{\mathbf{t}_i\}$ are to the ground-truth $\hat{\mathbf{t}}_i$---but we first need to account for the fundamental ambiguity (up to a similarity transform) in defining the world coordinate system.

To compute the transformation of camera translations under similarity transforms, let us consider a camera extrinsic $(R_i, \mathbf{t}_i)$ which maps world points $\mathbf{x}_w$ to camera frame $\mathbf{x}_c^i = R_i \mathbf{x}_w + \mathbf{t}_i$. 
If we defined a new world frame $\Tilde{\mathbf{x}}_w$ related to the current one via a similarity transform ${\bf x}_w = \frac{1}{s} (R \Tilde{\mathbf{x}}_w + \mathbf{t})$, and correspondingly define a scaled camera-frame coordinate system $\Tilde{\mathbf{x}}_c^i$ that follows $\mathbf{x}_c^i = \frac{1}{s} \Tilde{\mathbf{x}}_c^i$, we can see that the new extrinsics relating  $\Tilde{\mathbf{x}}_c^i$ to $\Tilde{\mathbf{x}}_w$ correspond to $\Tilde{R}_i = R_i R$, and $\Tilde{\mathbf{t}}_i = R_i \mathbf{t} + s \mathbf{t}_i$.
Therefore, to account for a similarity transform ambiguity, we search for the optimal $s, \mathbf{t}$ that minimize (using linear least-squares):
\begin{equation}
    \argmin_{s,\mathbf{t}} \sum_{i=1}^N \lVert \hat{\mathbf{t}}_i - (s\mathbf{t}_i  + R_i\mathbf{t})\rVert^2
\end{equation}

Note that  the above alignment objective does not care for the rotation component in world similarity transform. Intuitively, this is because the translation $\mathbf{t}_i$ corresponds to the coordinate of the world origin in the camera frame, and simply rotating the world coordinate system without moving its origin does not alter the origin's coordinate in camera frame. Given the optimal alignment, we measure the accuracy between the transformed translations $\Tilde{\mathbf{t}}_i = s\mathbf{t}_i + R_i\mathbf{t}$ and the GT translations $\hat{\mathbf{t}}_i$ as the the proportion of errors within 0.1 of the scene scale: $\lVert \hat{\mathbf{t}}_i - \Tilde{\mathbf{t}}_i \rVert < 0.1 \sigma$.

\subsection{Detailed Experiments}

We report the rotation and translation performances at additional thresholds, reporting joint rotation accuracy in \cref{tab:supp_rotation_thresholds} and camera center accuracy in \cref{tab:supp_cc_thresholds}.%

We report per-category evaluations for rotations in \cref{tab:supp_rotation_per_category} and camera centers in \cref{tab:supp_cc_per_category}. %
Among unseen object categories, we find that rotationally symmetric objects (ball, frisbee) are consistently challenging. Some categories may be ambiguous given few views (hotdog, kite), but these ambiguities are easy to resolve given more images. Objects that have many salient features (suitcase, remote) tend to have the highest performance.

\subsection{Additional Qualitative Results}

\begin{figure}
    \centering
    \includegraphics[width=\linewidth]{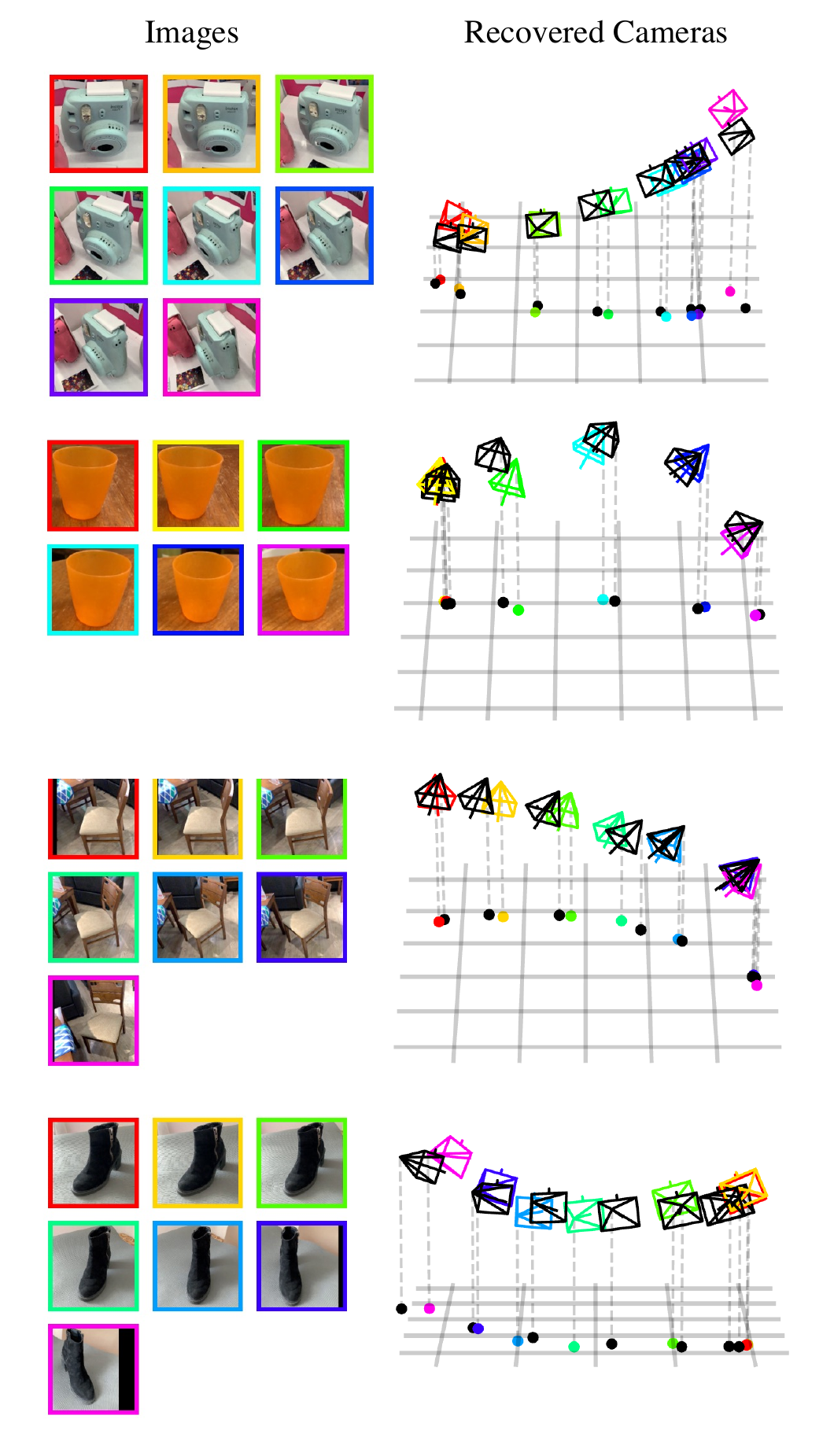}
    \caption{\textbf{Qualitative Results on the Objectron Dataset.} Here, we run our method on sequences from the Objectron dataset. Our method is trained only on CO3D without any fine-tuning, and we find that it generalizes well zero-shot. We visualize the ground truth camera poses in black and the predicted poses in color (with color corresponding to the input image).}
    \label{fig:supp_objectron}
\end{figure}

\begin{figure}
    \centering
    \includegraphics[width=\linewidth]{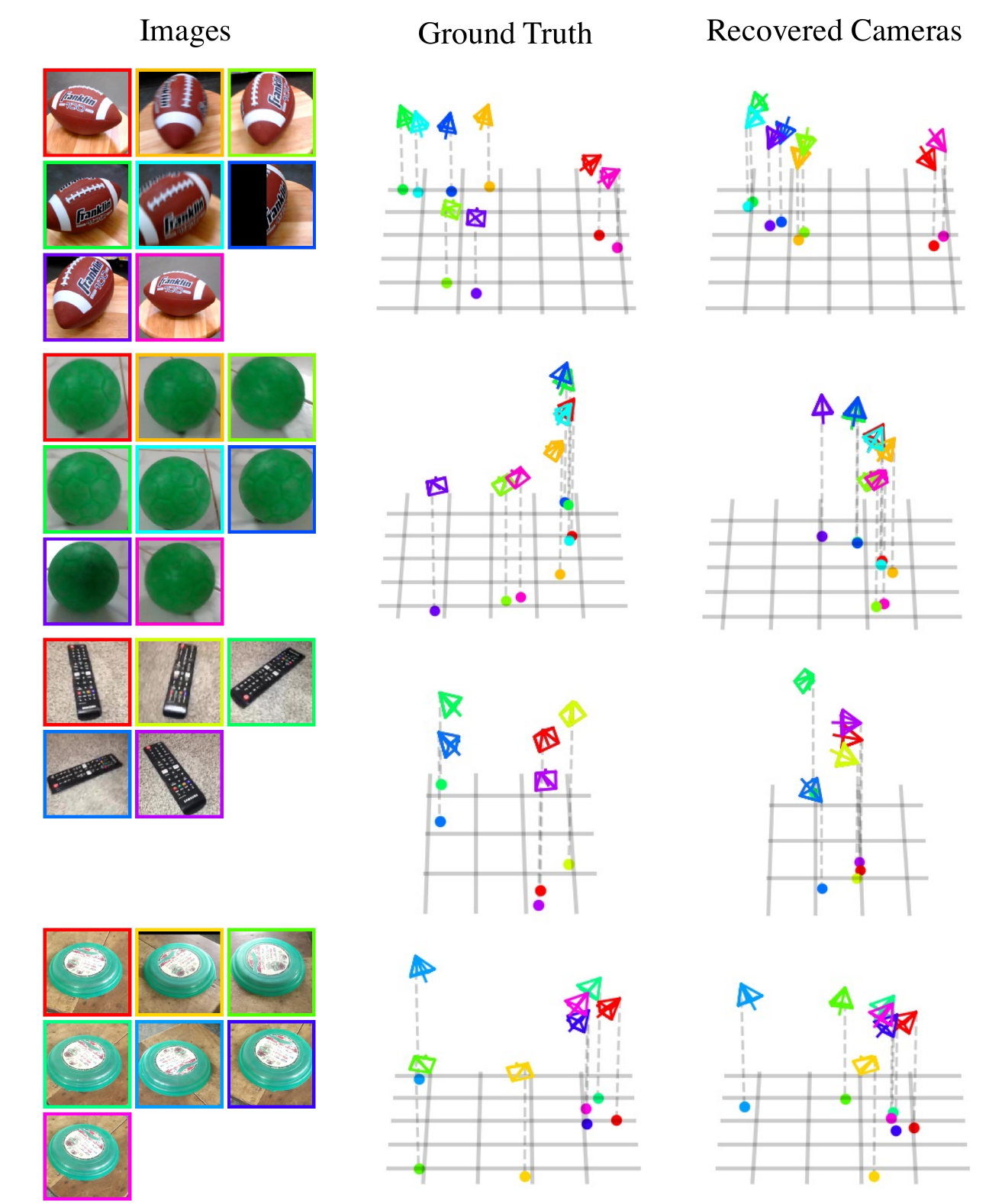}
    \caption{\textbf{Failure Modes.} Although handling symmetry is a primary motivation for our work, some symmetries are too challenging to handle. In the top two examples of the balls, our method cannot correctly recover the camera rotation since the object looks almost the same from multiple viewpoints. Unintuitively, our translation estimation is sometimes worse when all viewpoints are entirely fronto-parallel. This could be because the scene normalization procedure that places the world origin at the intersection of the optical axes at training time is unstable when all cameras are facing the exact same direction.}
    \label{fig:supp_failure}
\end{figure}

For discussion of failure modes, please refer to \cref{fig:supp_failure}. For qualitative results on the Objectron dataset, see \cref{fig:supp_objectron}.
For more qualitative results, please refer to the supplemental video.

\begin{figure}
    \centering
    \includegraphics[width=\linewidth]{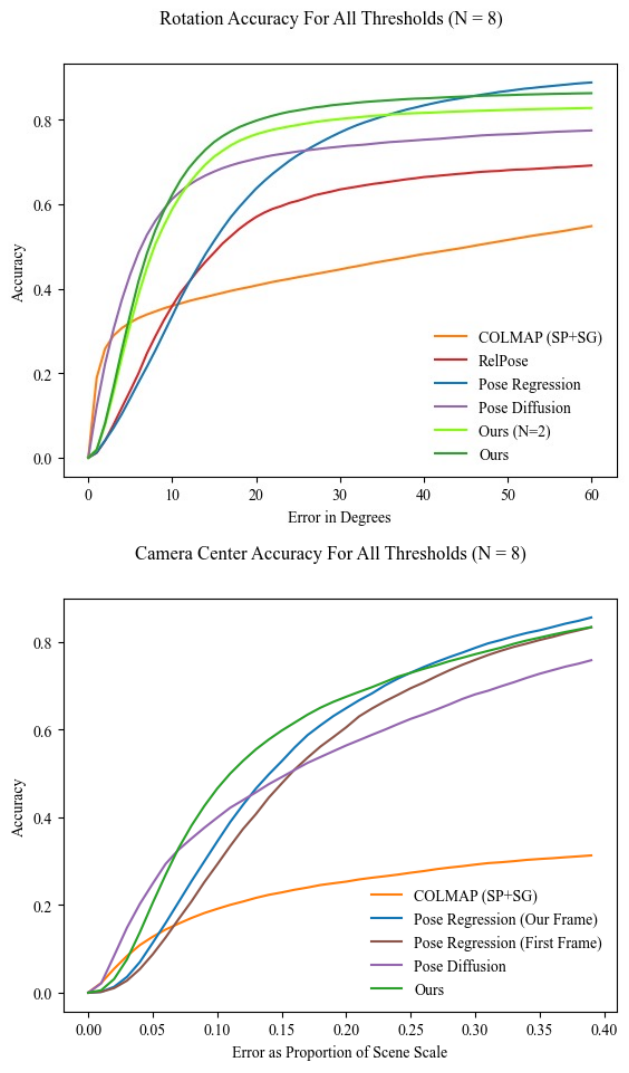}
    \caption{\textbf{Rotation and Camera Center Accuracy --- Area Under Curve.} We visualize the area under curve plots for rotation and camera center accuracy at fine error thresholds for 8 images of unseen categories. By visualizing very fine error thresholds ($\leq$5 degrees), we can see that COLMAP predicts highly accurate poses when it is able to converge. The concurrent PoseDiffusion is able to achieve slightly better localization for tighter thresholds, although our method quickly surpasses (7-8 degrees). We note that for downstream 3D reconstruction, approximate camera poses between 10-45 degrees of error are often sufficient using pose refinement methods such as BARF \cite{lin2021barf}.}
    \label{fig:auc}
\end{figure}

\begin{table}
\begin{center}
\footnotesize
\resizebox{\linewidth}{!}{
\begin{tabular}{llccccccc}
\toprule
& \# of Images & 2 & 3 & 4 & 5 & 6 & 7 & 8\\
\midrule
\parbox[t]{2mm}{\multirow{6}{*}{\rotatebox[origin=c]{90}{Seen Cate.}}} 
& COLMAP (SP+SG)~\cite{sarlin2019coarse} & 33.2 & 31.3 & 30.5 & 31.4 & 32.0 & 33.2 & 35.5 \\
&  RelPose~\cite{zhang2022relpose} & 59.5 & 61.2 & 61.2 & 61.1 & 61.1 & 61.2 & 61.1 \\
& PoseDiffusion~\cite{wang2023posediffusion} & 73.6 & 74.0 & 74.4 & 74.9 & 75.4 & 75.9 & 76.3 \\
\cmidrule(lr){2-9}
& Pose Regression  & 63.2 & 65.2 & 66.5 & 67.2 & 67.9 & 68.4 & 68.7\\
& Ours (N=2) & \textbf{78.2} & 78.8 & 79.0 & 79.3 & 79.3 & 79.5 & 79.5\\
& Ours (Full)   & \textbf{78.2} & \textbf{78.9} & \textbf{79.7} & \textbf{80.3} & \textbf{80.6} & \textbf{80.9} & \textbf{80.9} \\
\midrule
\parbox[t]{2mm}{\multirow{6}{*}{\rotatebox[origin=c]{90}{Unseen Cate.}}}
& COLMAP (SP+SG)~\cite{sarlin2019coarse} & 37.1 & 34.8 & 34.6 & 35.7 & 37.1 & 39.2 & 42.4 \\
& RelPose~\cite{zhang2022relpose} & 49.7 & 52.0 & 53.7 & 54.3 & 53.8 & 53.8 & 53.2\\
& PoseDiffusion~\cite{wang2023posediffusion} & 63.5 & 63.6 & 63.7 & 64.7 & 65.4 & 65.8 & 66.4 \\
\cmidrule(lr){2-9}
& Pose Regression & 56.8 & 59.1 & 60.2 & 61.5 & 62.3 & 62.8 & 63.5 \\
& Ours (N=2) & \textbf{67.6} & 68.5 & 68.9 & 68.6 & 68.7 & 68.9 & 69.3 \\
& Ours (Full)  & \textbf{67.6} & \textbf{69.5} & \textbf{70.3} & \textbf{71.3} & \textbf{72.0} & \textbf{72.2} & \textbf{72.5} \\
\bottomrule
\end{tabular}
}
\end{center}
\caption{\textbf{Joint Rotation Accuracy --- Area Under Curve\degree.} We evaluate rotation accuracy at every threshold by reporting the area under the curve from 0 to 60 degree error. See \cref{fig:auc} for a rotation accuracy curve visualized at fine error thresholds.
}
\label{tab:joint_rotation_auc}
\end{table}

\begin{table}
\begin{center}
\footnotesize
\resizebox{\linewidth}{!}{
\begin{tabular}{llccccccc}
\toprule
& \# of Images & 2 & 3 & 4 & 5 & 6 & 7 & 8\\
\midrule
\parbox[t]{2mm}{\multirow{5}{*}{\rotatebox[origin=c]{90}{Seen Cate.}}} 
& COLMAP (SP+SG)~\cite{sarlin2019coarse} & - & 33.8 & 24.0 & 19.5 & 16.8 & 16.1 & 16.9 \\
& PoseDiffusion~\cite{wang2023posediffusion} & - & 78.4 & 71.4 & 68.2 & 66.6 & 65.9 & 65.4 \\
\cmidrule(lr){2-9}
& Pose Reg. (First Fr.) & - & 74.5 & 65.8 & 62.2 & 60.4 & 59.3 & 58.5 \\
& Pose Reg. (Our Fr.) & - & 76.8 & 68.9 & 65.4 & 63.9 & 62.8 & 62.2 \\
&  Ours & - & \textbf{82.4} & \textbf{76.7} & \textbf{74.2} & \textbf{72.6} & \textbf{71.9} & \textbf{71.1} \\
\midrule
\parbox[t]{2mm}{\multirow{5}{*}{\rotatebox[origin=c]{90}{Unseen Cate.}}} 
& COLMAP (SP+SG)~\cite{sarlin2019coarse} & - & 35.6 & 26.4 & 21.7 & 20.1 & 20.4 & 22.2 \\
& PoseDiffusion~\cite{wang2023posediffusion} & - & 69.2 & 58.4 & 54.5 & 51.9 & 50.5 & 49.7 \\
\cmidrule(lr){2-9}
& Pose Reg. (First Fr.) & - & 67.8 & 58.2 & 54.2 & 52.2 & 50.7 & 49.7 \\
& Pose Reg. (Our Fr.) & - & 70.8 & 61.1 & 57.2 & 55.0 & 53.7 & 53.0 \\
&  Ours  & - & \textbf{72.8} & \textbf{64.4} & \textbf{60.4} & \textbf{58.5} & \textbf{56.9} & \textbf{56.3} \\
\bottomrule
\end{tabular}}
\end{center}
\caption{\textbf{Camera Center Accuracy --- Area Under Curve} We evaluate camera center accuracy at every threshold by reporting the area under the curve from 0 to .4 proportion of the scene scale. See \cref{fig:auc} for a rotation accuracy curve visualized at fine error thresholds.
}   
\label{tab:camera_center_auc}
\end{table}

\begin{table}
    \centering
    \footnotesize
    \begin{tabular}{ccccccccc}\toprule
    & \# of Image & 2 & 3 & 4 & 5 & 6 & 7 & 8\\
    \midrule
    \parbox[t]{2mm}{\multirow{4}{*}{\rotatebox[origin=c]{90}{Seen}}} 
    & Ours (Acc@5) &  42.1 & 43.6 & 44.4 & 44.7 & 45.0 & 45.1 & 45.4\\
    & Ours (Acc@10) &  70.5 & 72.3 & 73.6 & 74.2 & 74.7 & 75.2 & 75.2\\
    & Ours (Acc@15) &  81.8 & 82.8 & 84.1 & 84.7 & 84.9 & 85.3 & 85.5\\
    & Ours (Acc@30) &  89.2 & 90.1 & 91.0 & 91.5 & 91.7 & 92.0 & 92.1\\\cmidrule{2-9}
    \parbox[t]{2mm}{\multirow{4}{*}{\rotatebox[origin=c]{90}{Unseen}}} 
    & Ours (Acc@5) & 30.7 & 31.9 & 32.8 & 33.5 & 33.9 & 34.0 & 33.9\\
    & Ours (Acc@10) & 57.7 & 58.7 & 60.4 & 61.2 & 61.8 & 62.0 & 62.5\\
    & Ours (Acc@15) &  69.8 & 71.1 & 71.9 & 72.8 & 73.8 & 74.4 & 74.9\\
    & Ours (Acc@30) &  78.3 & 80.7 & 81.7 & 82.7 & 83.4 & 83.6 & 84.0\\
    \bottomrule
    \end{tabular}
    \caption{\textbf{Rotation accuracy at 5, 10, 15, and 30-degree thresholds.}}
    \label{tab:supp_rotation_thresholds}
\end{table}

\begin{table}
    \centering
    \footnotesize
    \begin{tabular}{ccccccccc}\toprule
    & \# of Image & 2 & 3 & 4 & 5 & 6 & 7 & 8\\
    \midrule
    \parbox[t]{2mm}{\multirow{3}{*}{\rotatebox[origin=c]{90}{Seen}}} 
    & Ours (Acc@0.1) & - & 85.0 & 78.0 & 74.2 & 71.9 & 70.3 & 68.8 \\
    & Ours (Acc@0.2) & - & 92.3 & 89.1 & 87.5 & 86.3 & 85.9 & 85.5 \\
    & Ours (Acc@0.3) & - & 95.4 & 92.8 & 91.5 & 90.7 & 90.8 & 90.4\\\cmidrule{1-9}
    \parbox[t]{2mm}{\multirow{3}{*}{\rotatebox[origin=c]{90}{Unseen}}} 
    & Ours (Acc@0.1) & - & 70.6 & 58.8 & 53.4 & 50.4 & 47.8 & 46.6\\
    & Ours (Acc@0.2) &  - & 82.5 & 75.6 & 71.9 & 69.9 & 68.5 & 67.5\\
    & Ours (Acc@0.3) & - & 88.7 & 83.1 & 80.3 & 78.8 & 77.6 & 77.2\\
    \bottomrule
    \end{tabular}
    \caption{\textbf{Camera center evaluation at 0.1, 0.2, 0.3 thresholds.}}
    \label{tab:supp_cc_thresholds}
\end{table}

\begin{table}
    \centering
    \footnotesize
    \begin{tabular}{ccccccccc}\toprule
    & \# of Image & 2 & 3 & 4 & 5 & 6 & 7 & 8\\
    \midrule
    \parbox[t]{2mm}{\multirow{41}{*}{\rotatebox[origin=c]{90}{Seen Categories}}}
    & apple & 74.8 & 75.1 & 75.3 & 77.6 & 78.5 & 79.3 & 79.5 \\
& backpack & 85.0 & 87.3 & 88.4 & 89.0 & 89.7 & 89.9 & 89.8 \\
& banana & 84.1 & 83.9 & 84.8 & 85.1 & 86.0 & 86.4 & 86.6 \\
& baseballbat & 84.3 & 83.3 & 89.5 & 90.7 & 89.8 & 90.7 & 91.1 \\
& baseballglove & 76.0 & 69.8 & 73.8 & 75.7 & 74.5 & 77.3 & 77.4 \\
& bench & 89.6 & 88.9 & 89.9 & 89.5 & 90.1 & 90.1 & 89.6 \\
& bicycle & 81.6 & 85.9 & 87.1 & 88.4 & 88.0 & 87.8 & 87.9 \\
& bottle & 77.6 & 81.2 & 82.3 & 82.4 & 83.8 & 83.6 & 84.1 \\
& bowl & 88.7 & 90.2 & 90.2 & 91.0 & 90.6 & 91.2 & 91.3 \\
& broccoli & 62.5 & 64.6 & 69.6 & 68.5 & 69.8 & 69.6 & 70.9 \\
& cake & 75.6 & 78.4 & 78.7 & 78.4 & 78.5 & 78.4 & 78.8 \\
& car & 86.8 & 87.4 & 89.6 & 90.9 & 91.4 & 90.5 & 90.8 \\
& carrot & 83.7 & 84.9 & 86.5 & 86.1 & 86.5 & 86.6 & 86.4 \\
& cellphone & 85.6 & 85.9 & 86.6 & 86.8 & 86.3 & 87.2 & 87.6 \\
& chair & 94.9 & 97.0 & 97.5 & 97.6 & 98.1 & 98.6 & 98.4 \\
& cup & 69.2 & 71.2 & 72.4 & 73.1 & 74.3 & 74.4 & 75.6 \\
& donut & 59.2 & 62.7 & 60.9 & 64.1 & 65.3 & 65.8 & 66.0 \\
& hairdryer & 86.1 & 86.5 & 86.5 & 88.2 & 88.6 & 89.1 & 88.8 \\
& handbag & 77.8 & 80.4 & 82.1 & 82.7 & 82.2 & 82.3 & 82.7 \\
& hydrant & 94.0 & 92.5 & 94.5 & 95.4 & 96.3 & 95.6 & 95.7 \\
& keyboard & 89.4 & 90.8 & 92.4 & 92.8 & 93.1 & 92.9 & 92.9 \\
& laptop & 96.8 & 96.9 & 97.2 & 97.3 & 97.4 & 97.1 & 97.0 \\
& microwave & 78.4 & 81.3 & 81.9 & 81.9 & 81.2 & 82.8 & 81.3 \\
& motorcycle & 85.2 & 88.9 & 88.9 & 90.2 & 90.3 & 90.2 & 89.7 \\
& mouse & 89.0 & 89.7 & 90.5 & 90.5 & 90.7 & 90.8 & 90.8 \\
& orange & 70.9 & 70.9 & 72.1 & 73.4 & 74.0 & 75.4 & 74.9 \\
& parkingmeter & 70.0 & 65.6 & 65.6 & 67.0 & 66.9 & 67.9 & 69.9 \\
& pizza & 78.1 & 80.3 & 83.8 & 83.1 & 84.1 & 85.5 & 84.6 \\
& plant & 73.8 & 74.6 & 74.7 & 75.9 & 75.8 & 77.2 & 77.3 \\
& stopsign & 84.5 & 87.8 & 88.6 & 88.8 & 89.2 & 88.6 & 89.2 \\
& teddybear & 85.0 & 86.3 & 86.4 & 87.7 & 88.3 & 88.5 & 88.3 \\
& toaster & 90.4 & 93.9 & 94.3 & 95.2 & 94.7 & 95.6 & 95.4 \\
& toilet & 93.1 & 94.0 & 95.1 & 95.7 & 96.1 & 95.8 & 96.6 \\
& toybus & 88.5 & 86.4 & 86.0 & 85.9 & 84.7 & 86.0 & 85.6 \\
& toyplane & 64.1 & 68.4 & 70.5 & 70.5 & 70.3 & 71.0 & 71.9 \\
& toytrain & 82.5 & 86.5 & 88.2 & 87.1 & 87.4 & 88.5 & 88.8 \\
& toytruck & 81.3 & 82.4 & 85.1 & 85.9 & 86.5 & 86.5 & 87.8 \\
& tv & 100.0 & 100.0 & 100.0 & 100.0 & 100.0 & 100.0 & 100.0 \\
& umbrella & 81.6 & 84.5 & 85.3 & 85.0 & 85.8 & 86.8 & 86.4 \\
& vase & 77.4 & 79.6 & 80.4 & 81.3 & 81.3 & 82.5 & 81.9 \\
& wineglass & 72.6 & 70.9 & 72.9 & 74.3 & 74.3 & 73.8 & 74.4 \\
& Mean & 81.8 & 82.8 & 84.1 & 84.7 & 84.9 & 85.3 & 85.5 \\
\cmidrule{1-9}
\parbox[t]{2mm}{\multirow{10}{*}{\rotatebox[origin=c]{90}{Unseen Categories}}} 
& ball & 52.5 & 53.9 & 54.7 & 55.2 & 55.3 & 56.3 & 56.1 \\
& book & 71.1 & 72.7 & 76.0 & 77.2 & 78.4 & 77.2 & 77.9 \\
& couch & 80.0 & 81.7 & 82.6 & 82.2 & 83.1 & 84.8 & 85.3 \\
& frisbee & 67.2 & 69.3 & 71.7 & 74.2 & 75.8 & 74.1 & 76.3 \\
& hotdog & 61.4 & 59.0 & 58.3 & 58.1 & 61.1 & 65.6 & 66.3 \\
& kite & 63.1 & 66.4 & 67.8 & 70.3 & 69.4 & 69.2 & 69.8 \\
& remote & 74.4 & 78.1 & 81.6 & 82.2 & 82.2 & 83.4 & 84.1 \\
& sandwich & 66.0 & 67.7 & 68.3 & 67.5 & 67.8 & 69.9 & 68.9 \\
& skateboard & 71.1 & 71.1 & 67.0 & 69.1 & 72.4 & 71.9 & 71.7 \\
& suitcase & 91.6 & 90.9 & 91.2 & 92.4 & 92.7 & 91.8 & 92.4 \\
& Mean  & 69.8 & 71.1 & 71.9 & 72.8 & 73.8 & 74.4 & 74.9\\
    \bottomrule
    \end{tabular}
    \caption{\textbf{Per category rotation accuracy at 15 degrees.}}
    \label{tab:supp_rotation_per_category}
\end{table}

\begin{table}
    \centering
    \footnotesize
    \begin{tabular}{ccccccccc}\toprule
    & \# of Image & 2 & 3 & 4 & 5 & 6 & 7 & 8\\
    \midrule
    \parbox[t]{2mm}{\multirow{41}{*}{\rotatebox[origin=c]{90}{Seen Categories}}}
    & apple & - & 96.1 & 94.6 & 94.1 & 93.5 & 91.4 & 91.8 \\
& backpack & - & 93.6 & 90.3 & 90.1 & 89.1 & 89.0 & 88.9 \\
& banana & - & 98.1 & 94.2 & 93.5 & 92.3 & 91.0 & 90.8 \\
& baseballbat & - & 88.1 & 83.6 & 84.9 & 80.2 & 79.4 & 82.0 \\
& baseballglove & - & 91.1 & 85.7 & 83.5 & 80.4 & 82.1 & 80.3 \\
& bench & - & 91.1 & 86.2 & 85.7 & 86.3 & 85.1 & 82.9 \\
& bicycle & - & 91.5 & 89.5 & 89.0 & 87.7 & 87.6 & 86.4 \\
& bottle & - & 93.5 & 89.6 & 87.0 & 86.8 & 86.5 & 84.9 \\
& bowl & - & 94.0 & 91.5 & 91.0 & 90.5 & 90.0 & 90.5 \\
& broccoli & - & 94.4 & 92.8 & 91.1 & 89.8 & 89.2 & 89.7 \\
& cake & - & 89.6 & 84.0 & 81.6 & 80.1 & 79.6 & 80.1 \\
& car & - & 91.9 & 92.2 & 92.8 & 89.3 & 90.2 & 89.7 \\
& carrot & - & 95.1 & 92.9 & 91.9 & 90.1 & 89.4 & 88.0 \\
& cellphone & - & 92.3 & 90.2 & 85.2 & 82.9 & 81.7 & 82.6 \\
& chair & - & 98.6 & 98.2 & 97.9 & 97.7 & 98.4 & 98.2 \\
& cup & - & 89.6 & 83.6 & 81.5 & 80.5 & 79.4 & 80.4 \\
& donut & - & 81.9 & 74.8 & 70.2 & 70.1 & 68.1 & 69.4 \\
& hairdryer & - & 98.1 & 96.4 & 95.3 & 94.6 & 94.8 & 94.4 \\
& handbag & - & 88.6 & 84.4 & 82.3 & 80.7 & 78.8 & 77.9 \\
& hydrant & - & 95.6 & 96.6 & 96.0 & 96.3 & 95.7 & 95.4 \\
& keyboard & - & 92.7 & 90.7 & 90.5 & 89.2 & 87.6 & 86.9 \\
& laptop & - & 98.5 & 96.5 & 94.6 & 93.5 & 93.7 & 92.8 \\
& microwave & - & 85.1 & 79.2 & 74.7 & 69.7 & 74.2 & 71.9 \\
& motorcycle & - & 98.1 & 96.4 & 95.4 & 96.2 & 95.3 & 94.2 \\
& mouse & - & 98.6 & 97.6 & 97.8 & 97.3 & 97.2 & 95.9 \\
& orange & - & 92.0 & 87.4 & 84.3 & 84.3 & 83.4 & 83.3 \\
& parkingmeter & - & 83.3 & 71.7 & 70.0 & 70.0 & 67.6 & 72.5 \\
& pizza & - & 89.2 & 88.6 & 84.8 & 85.2 & 84.6 & 82.7 \\
& plant & - & 91.4 & 84.9 & 82.6 & 83.4 & 83.4 & 82.5 \\
& stopsign & - & 95.6 & 91.5 & 90.4 & 89.3 & 86.5 & 86.6 \\
& teddybear & - & 96.3 & 95.1 & 94.1 & 94.6 & 94.2 & 93.2 \\
& toaster & - & 94.9 & 93.5 & 94.0 & 92.2 & 94.1 & 93.5 \\
& toilet & - & 94.7 & 89.3 & 88.0 & 85.2 & 84.5 & 84.7 \\
& toybus & - & 90.5 & 86.5 & 84.3 & 81.2 & 86.0 & 83.1 \\
& toyplane & - & 86.3 & 82.1 & 78.6 & 75.4 & 74.8 & 74.2 \\
& toytrain & - & 88.3 & 87.2 & 84.6 & 83.8 & 82.4 & 82.2 \\
& toytruck & - & 88.4 & 83.9 & 81.3 & 80.4 & 78.9 & 79.7 \\
& tv & - & 95.6 & 100.0 & 100.0 & 100.0 & 100.0 & 100.0 \\
& umbrella & - & 95.5 & 94.5 & 93.0 & 92.0 & 92.6 & 90.6 \\
& vase & - & 90.4 & 85.2 & 83.6 & 82.3 & 81.7 & 80.5 \\
& wineglass & - & 85.8 & 78.1 & 75.9 & 72.6 & 71.6 & 71.6 \\
& Mean & - & 92.3 & 89.1 & 87.5 & 86.3 & 85.9 & 85.5\\
\cmidrule{2-9}
\parbox[t]{2mm}{\multirow{10}{*}{\rotatebox[origin=c]{90}{Unseen Categories}}} 
& ball & - & 85.4 & 77.1 & 72.9 & 69.4 & 67.7 & 65.2 \\
& book & - & 78.9 & 72.1 & 70.5 & 68.8 & 67.1 & 66.2 \\
& couch & - & 85.9 & 78.5 & 75.3 & 72.5 & 71.8 & 71.9 \\
& frisbee & - & 82.1 & 80.2 & 77.4 & 78.0 & 76.1 & 76.1 \\
& hotdog & - & 70.0 & 61.4 & 56.3 & 54.8 & 56.3 & 56.2 \\
& kite & - & 82.6 & 71.9 & 66.0 & 60.5 & 57.7 & 57.1 \\
& remote & - & 89.2 & 82.9 & 79.8 & 79.5 & 78.5 & 78.9 \\
& sandwich & - & 83.5 & 77.6 & 72.1 & 68.3 & 67.6 & 66.8 \\
& skateboard & - & 75.9 & 65.3 & 58.4 & 58.0 & 55.7 & 50.0 \\
& suitcase & - & 92.0 & 89.1 & 90.1 & 89.3 & 87.0 & 87.0 \\
& Mean & - & 82.5 & 75.6 & 71.9 & 69.9 & 68.5 & 67.5 \\
    \bottomrule
    \end{tabular}
    \caption{\textbf{Per category camera center accuracy at 0.2}}
    \label{tab:supp_cc_per_category}
\end{table}

\end{document}